\documentclass{article}
\PassOptionsToPackage{numbers}{natbib}
\usepackage[preprint]{neurips_2026}

\usepackage[utf8]{inputenc}
\usepackage[T1]{fontenc}
\usepackage{amsmath,amssymb}
\usepackage{amsfonts}
\usepackage{textcomp}   
\usepackage{booktabs}
\usepackage{titletoc}
\usepackage{multirow}
\usepackage{flafter}
\usepackage{graphicx}
\usepackage{capt-of}    
\usepackage{placeins}
\usepackage{float}
\usepackage{wrapfig}
\usepackage{needspace}
\usepackage{url}
\usepackage{makecell}
\usepackage{pifont}
\usepackage{microtype}
\usepackage{nicefrac}
\usepackage{subcaption}
\usepackage[table]{xcolor}
\usepackage[most]{tcolorbox}
\usepackage{listings}
\usepackage{hyperref}
\hypersetup{hypertexnames=false}
\usepackage[nameinlink,capitalise,noabbrev]{cleveref}

\definecolor{bestfill}{RGB}{157,195,230}   
\definecolor{secondfill}{RGB}{204,224,243} 
\definecolor{thirdfill}{RGB}{236,244,252}  
\definecolor{sepgray}{gray}{0.45}
\definecolor{promptbg}{RGB}{242,242,242}
\definecolor{promptborder}{RGB}{128,128,128}
\lstdefinestyle{prompttext}{
  basicstyle=\ttfamily\fontsize{7.4}{9.1}\selectfont,
  breaklines=true,
  breakatwhitespace=true,
  breakautoindent=false,
  breakindent=0pt,
  columns=spaceflexible,
  keepspaces=true,
  showstringspaces=false,
  aboveskip=0pt,
  belowskip=0pt,
  upquote=false,
  literate={“}{{\textnormal{``}}}1 {”}{{\textnormal{''}}}1 {‘}{{\textnormal{`}}}1 {’}{{\textnormal{'}}}1
}
\newcommand{\promptinput}[1]{%
  \tcbinputlisting{
    enhanced,
    breakable,
    listing only,
    listing file={#1},
    listing options={style=prompttext},
    colback=promptbg,
    colframe=promptborder,
    boxrule=0.4pt,
    toprule at break=0pt,
    bottomrule at break=0pt,
    arc=2pt,
    boxsep=0pt,
    left=7pt,
    right=7pt,
    top=4pt,
    bottom=4pt,
    before skip=6pt,
    after skip=12pt
  }%
}
\newsavebox{\hlvalbox}
\newsavebox{\hlvalpatch}
\newdimen\hlvalpad
\newdimen\hlvalvpad
\newdimen\hlvalwidth
\makeatletter
\newcommand{\hlcell}[2]{%
  \begingroup
  \sbox\hlvalbox{#2}%
  \hlvalwidth=\dimexpr\wd\hlvalbox + 2\hlvalpad\relax
  \sbox\hlvalpatch{%
    \hbox to \hlvalwidth{%
      \rlap{\color{#1}\vrule width \hlvalwidth
             height \dimexpr\ht\@arstrutbox + \hlvalvpad\relax
             depth  \dimexpr\dp\@arstrutbox + \hlvalvpad\relax}%
      \hss\box\hlvalbox\hss}}%
  \leavevmode\box\hlvalpatch
  \endgroup}
\makeatother
\newdimen\hlcellw
\newcommand{\hlhead}[1]{\makebox[\hlcellw][c]{#1}}
\makeatletter
\newcommand{\hlcellwide}[2]{%
  \begingroup
  \sbox\hlvalbox{#2}%
  \sbox\hlvalpatch{%
    \hbox to \hlcellw{%
      \rlap{\hspace*{-\tabcolsep}\color{#1}\vrule
             width \dimexpr\hlcellw + 2\tabcolsep\relax
             height \dimexpr\ht\@arstrutbox + \hlvalvpad\relax
             depth  \dimexpr\dp\@arstrutbox + \hlvalvpad\relax}%
      \hss\box\hlvalbox\hss}}%
  \leavevmode\box\hlvalpatch
  \endgroup}
\makeatother
\usepackage{tikz}
\usetikzlibrary{positioning, arrows.meta}

\newcommand{\benchmark}{LabInstruct}
\newcommand{\numdisciplines}{5}
\newcommand{\numtasks}{204}
\newcommand{\numlevelone}{101}
\newcommand{\numleveltwo}{103}
\newcommand{\nummodels}{8}
\newcommand{\cmark}{$\checkmark$}
\newcommand{\xmark}{$\times$}
\newcommand{\colhead}[1]{\begin{tabular}[c]{@{}c@{}}#1\end{tabular}}

\newcommand{\colorboxinline}[1]{\textcolor{myblue}{\rule{0.8em}{0.8em}}}
\usepackage[utf8]{inputenc}
\usepackage{newunicodechar}
\newunicodechar{♫}{\ding{14}} 

\title{\texorpdfstring{\benchmark: Benchmarking Situated Instructional\\Video Generation for Lab Procedures}{LabInstruct: Benchmarking Situated Instructional Video Generation for Lab Procedures}}

\author{%
  \normalfont
  \textbf{Yuming Fu}$^{1,*}$, \textbf{Weijia Wu}$^{2,*}$, \textbf{Jing Chen}$^{3,*}$,
  \textbf{Jiahao Tang}$^{1}$, \textbf{Feifei Chen}$^{1}$, \textbf{Hongyu Zhu}$^{4}$,\\
  \textbf{Xin Jin}$^{5}$, \textbf{Alex Jinpeng Wang}$^{1,\dagger}$ \\[2pt]
  $^{1}$Central South University \quad
  $^{2}$National University of Singapore \\
  $^{3}$Beijing University of Posts and Telecommunications \quad
  $^{5}$Monash University \\
  $^{4}$Chongqing Institute of Green and Intelligent Technology, Chinese Academy of Sciences
}

\definecolor{ncmetabg}{HTML}{E9EEF5}
\definecolor{ncmetaedge}{HTML}{DCE6F5}
\definecolor{ncmetablue}{HTML}{1877F2}
\definecolor{ncapricot}{HTML}{F7EBDD}

\makeatletter
\newcommand{\maketitleboxed}[1]{%
  \par
  \begingroup
    \renewcommand{\thefootnote}{\fnsymbol{footnote}}
    \renewcommand{\@makefnmark}{\hbox to \z@{$^{\@thefnmark}$\hss}}
    \long\def\@makefntext##1{%
      \parindent 1em\noindent
      \hbox to 1.8em{\hss $\m@th ^{\@thefnmark}$}##1%
    }
    \thispagestyle{empty}%
    \begin{tcolorbox}[
      enhanced,
      colback=ncmetabg,
      colframe=ncmetaedge,
      boxrule=0.35pt,
      arc=12pt,
      left=0.55cm, right=0.55cm, top=0.45cm, bottom=0.4cm,
      before skip=0pt, after skip=0.4em,
      grow to left by=1.5pt, grow to right by=1.5pt,
    ]
      \centering
      {\Large\bf \raisebox{-0.32\height}{\includegraphics[height=2.4em]{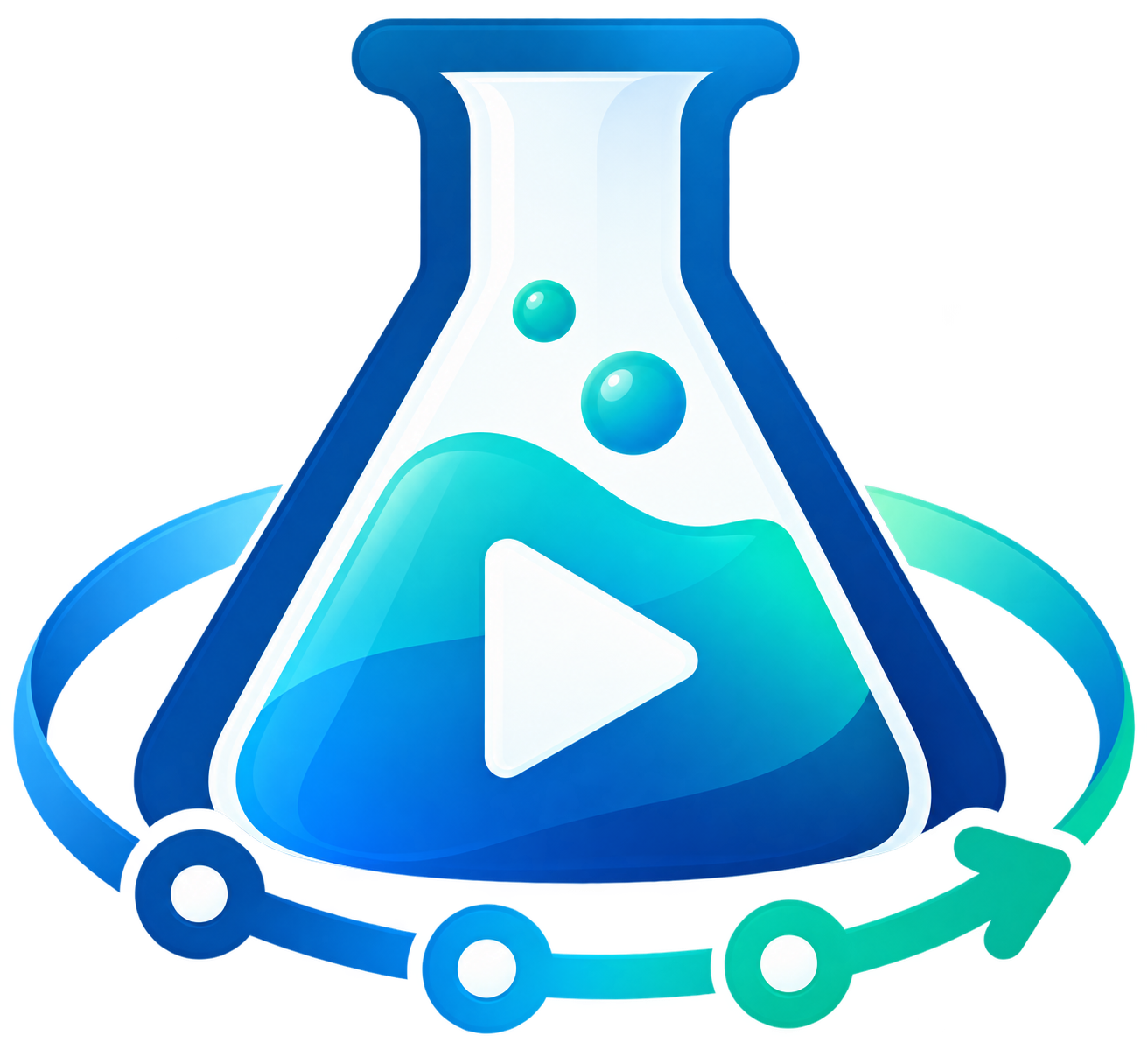}}\,\@title\par}%
      \vskip 0.22in
      \def\And{\end{tabular}\hfil\linebreak[0]\hfil\begin{tabular}[t]{c}\bf\rule{\z@}{24\p@}\ignorespaces}%
      \def\AND{\end{tabular}\hfil\linebreak[4]\hfil\begin{tabular}[t]{c}\bf\rule{\z@}{24\p@}\ignorespaces}%
      \begin{tabular}[t]{c}\bf\rule{\z@}{24\p@}\@author\end{tabular}\par
      \vskip 0.18in
      \begingroup
        \leftskip=1.5em \rightskip=1.5em
        \centerline{\large\bf Abstract}\vspace{0.6ex}
        \small #1\par
        \vspace{0.6ex}
        {\parfillskip=0pt plus 1fil\relax
         \noindent\small\textbf{Project page:~}~\href{https://csu-jpg.github.io/LabInstruct.github.io/}{\color{magenta}\texttt{https://csu-jpg.github.io/LabInstruct.github.io/}}\par}
      \endgroup
    \end{tcolorbox}%
    \@thanks
  \endgroup
  \let\maketitle\relax
  \let\thanks\relax
}
\makeatother

\begin{document}

\maketitleboxed{Self-driving laboratories (SDLs) aim to automate the full experimental loop, from scientific decision-making to physical execution. Ideally, AI-generated plans could be carried out directly by robotic systems, but reliable automation remains difficult in complex, open-world laboratory environments, where experiments often involve fine-grained manipulation, long-horizon procedures, and substantial variation across tasks and setups. Humans therefore remain an important execution interface between AI-generated plans and physical experiments, creating a need for clear and effective human-facing experimental guidance. Because laboratory procedures are inherently visual, spatial, and dynamic, video is particularly well suited to communicating apparatus configurations, manipulation actions, temporal dependencies, and state changes. Recent advances in video generation now make it possible to synthesize experimental demonstrations directly from an initial workspace image and a natural-language instruction. However, whether such models can reliably communicate real laboratory procedures has not been systematically studied. We introduce \textbf{\benchmark}, a benchmark for situated instructional video generation in real laboratories. \benchmark\ contains 204 tasks across 5 scientific disciplines, with real reference executions and structured annotations of objects, actions, contacts, and state transitions. Evaluating 8 frontier image-to-video models, we find that visually plausible generations frequently remain procedurally incorrect, revealing a substantial gap between visual realism and the reliability required for experimental instruction.
}

{\makeatletter
 \renewcommand{\thefootnote}{0}%
 \long\def\@makefntext#1{\parindent 1em\noindent \hb@xt@1.8em{\hss}#1}%
 \makeatother
 \footnotetext[0]{\textsuperscript{*}Equal contribution.\quad\textsuperscript{\textdagger}Corresponding author.}}

\begin{figure}[!ht]
    \centering
    \includegraphics[width=0.98\linewidth]{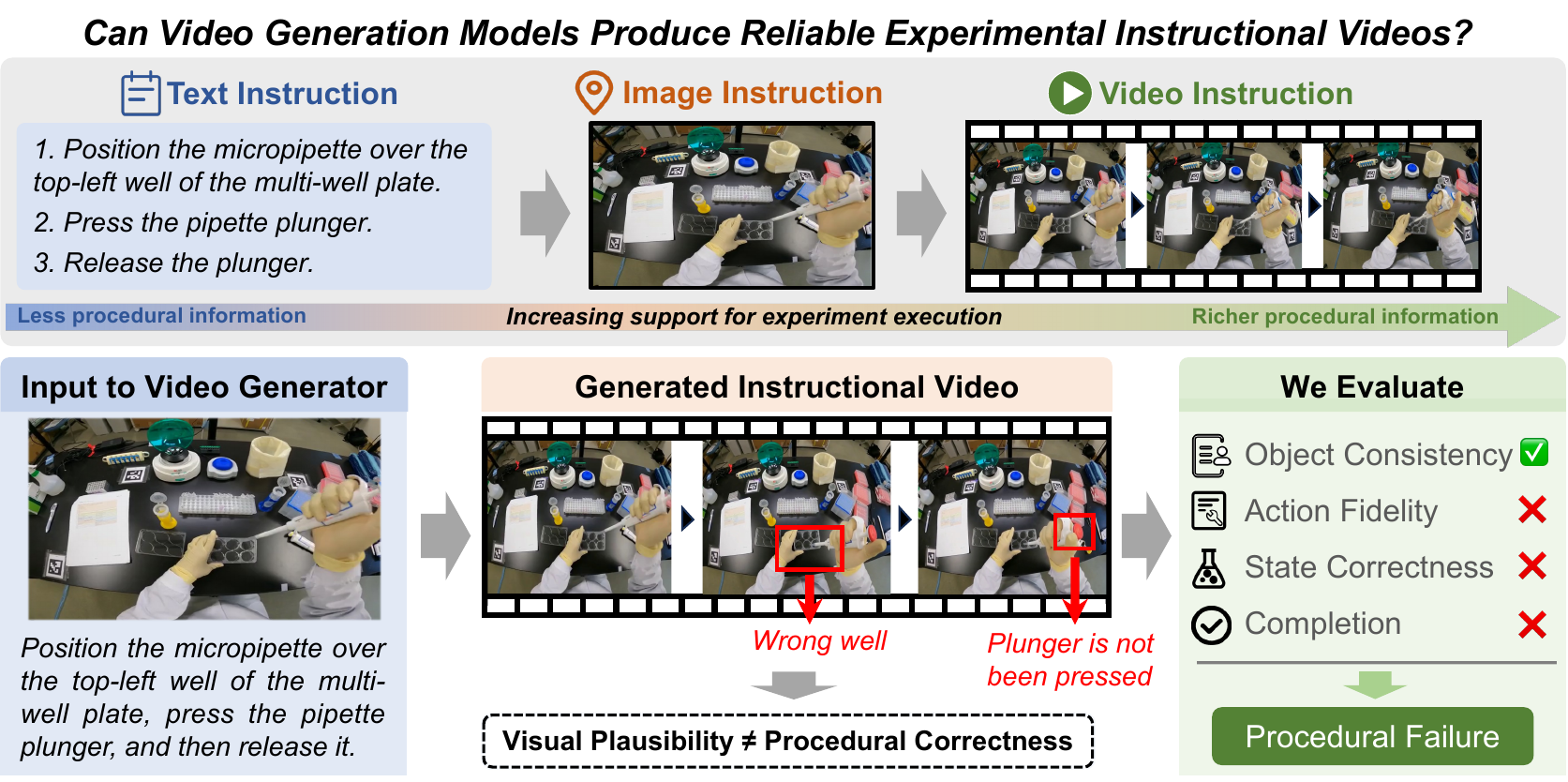}
    \caption{\textbf{Overview of \benchmark.} Text, image, and video provide increasingly rich procedural guidance, while \benchmark\ evaluates generated videos for procedural correctness beyond visual plausibility.}
    \label{fig:benchmark_overview}
\end{figure}

\section{Introduction}

Self-driving laboratories (SDLs) seek to accelerate discovery by integrating AI, automated experimentation, and iterative decision-making \cite{szymanski2023autonomous,zhang2025multimodal}. Recent systems support hypothesis generation and scientific software development \cite{gottweis2026coscientist,ghareeb2026multiagent,aygun2026era}, multi-step decision-making \cite{ferber2026medical}, and experiment planning, execution, and closed-loop optimization \cite{boiko2023autonomous,bennett2026autonomous}. Yet reliable physical execution remains difficult in real laboratories, where diverse instruments, fine-grained manipulations, long action sequences, and workspace-specific constraints challenge embodied systems. Accordingly, current laboratory-robot and scientific-agent benchmarks often focus on restricted simulations or software workflows \cite{lan2026autobio,sun2026scienceboard}. Humans therefore remain a practical interface between AI-generated plans and physical experiments, making it essential to communicate not only \emph{what} to do, but also \emph{how} to do it.

Experimental instruction is thus a key component of human-in-the-loop laboratory automation. While text can specify goals and symbolic steps, laboratory operations are inherently visual, spatial, and dynamic, involving apparatus placement, tool--object contacts, motion, intermediate states, and temporal order that text alone may leave ambiguous. Images and demonstration videos can ground instructions in the current workspace and make required actions and state changes directly observable. Prior work has explored this direction through action and state-change image synthesis, step-by-step visual instructions, and language-conditioned video continuation from observed scenes \cite{soucek2024genhowto,soucek2025showhowto,gu2024seer,xing2025aid,pu2026showme}.

Recent video generation models can synthesize demonstrations from workspace images and instructions, potentially bridging AI-driven experimentation and human execution (Figure~\ref{fig:benchmark_overview}). Yet visual realism does not ensure reliable guidance. For instructional use, generation quality depends not only on visual plausibility, but also on whether the depicted actions correctly realize the intended procedure. Existing benchmarks evaluate visual fidelity~\cite{huang2024vbench,zheng2025vbench2}, prompt alignment~\cite{li2026worldmodelbench}, physical plausibility~\cite{motamed2026physicsiq,zhang2025morpheus}, scientific reasoning~\cite{hu2025videoscience,feng2026rewardmap,feng2026reasonmap}, or educational explanation~\cite{mariam2026phyeduvideo}, while laboratory and egocentric datasets focus on recognition, parsing, reasoning, and evidence localization in recorded videos \cite{nishimura2021biovl,yagi2025finebio,zou2025physlab,xu2025expvid,ragusa2026egoextra,nagrani2026minervaego}. None evaluates whether a model can correctly generate human-executed procedures in real laboratories.

We therefore ask: \textit{\textbf{Can video generation models produce reliable experimental instructional videos?}} Such videos must use the required apparatus, act on the correct objects, preserve causal order, produce intended state transitions, and visibly reach the outcome. A visually plausible generation may omit critical actions, manipulate the wrong object, reorder dependent steps, or terminate prematurely. Conversely, a valid execution need not reproduce the reference trajectory exactly, as long as it preserves the workspace and satisfies the procedural requirements. Experimental instruction therefore imposes stricter demands than conventional visual synthesis, extending language-guided laboratory manipulation \cite{lan2026autobio} from embodied agents to human-facing video.

\begin{table}[t]
    \centering
    \footnotesize
    \caption{\textbf{Comparison of scientific video generation benchmarks.} Lab denotes benchmarks based on real laboratory scenes, Ref. indicates the availability of reference executions, and Proc. denotes human-executed procedures.}
    \label{tab:benchmark_comparison}
    \begingroup
    \setlength{\tabcolsep}{8pt}
    \renewcommand{\arraystretch}{1}
    \resizebox{\linewidth}{!}{%
    \begin{tabular}{@{}lcccccccc@{}}
        \toprule
        \textbf{Benchmark} & \textbf{Year} & \textbf{Publication} & \textbf{\#Prompts} & \textbf{Discipline} & \textbf{Paradigm} & \textbf{Lab} & \textbf{Ref.} & \textbf{Proc.} \\
        \midrule
        VideoPhy-2~\citep{bansal2025videophy2} & 2025 & ICLR & 688 & Physics & T2V & \xmark & \xmark & \xmark \\
        VideoScience-Bench~\citep{hu2025videoscience} & 2025 & arXiv & 200 & Phys./Chem. & T2V/I2V & \xmark & \cmark & \xmark \\
        Physics-IQ~\citep{motamed2026physicsiq} & 2026 & WACV & 66 & Physics & I2V & \xmark & \cmark & \xmark \\
        PhyGround~\citep{lin2026phyground} & 2026 & arXiv & 250 & Physics & TI2V & \xmark & \xmark & \xmark \\
        Morpheus~\citep{zhang2025morpheus} & 2026 & ICML & 130 & Physics & I2V & \xmark & \cmark & \xmark \\
        Sci-VBench~\citep{zhang2026scivbench} & 2026 & COLM & 1,253 & Multi. & T2V & \xmark & \xmark & \xmark \\
        \textbf{\benchmark} & \textbf{2026} & \textbf{--} & \textbf{\numtasks} & \textbf{Multi.} & \textbf{TI2V} & \cmark & \cmark & \cmark \\
        \bottomrule
    \end{tabular}}
    \endgroup
\end{table}

We introduce \benchmark, a benchmark for situated instructional video generation in real laboratories. Given an initial workspace observation and a natural-language instruction, a model must generate a continuation that correctly demonstrates the procedure. \benchmark\ contains 204 tasks across 5 scientific disciplines and two complexity levels, from single experimental steps to short multi-step procedures with inter-step dependencies. Each task includes a real reference execution and structured annotations of objects, actions, contacts, and state transitions. Our contributions are: \textbf{i)} formulating laboratory video generation as a human-facing procedural communication task; \textbf{ii)} introducing \benchmark, a real-world benchmark with reference executions and structured procedural annotations; and \textbf{iii)} developing a procedure-aware evaluation and benchmarking 8 frontier image-to-video models to quantify the gap between visual plausibility and end-to-end procedural correctness under realistic laboratory conditions.

\section{Related Work}

\textbf{Visual Instruction Generation.} Visual instruction generation shows how to perform an action or change an object under a language command. GenHowTo~\cite{soucek2024genhowto} generates action and state-change images, while ShowHowTo~\cite{soucek2025showhowto} extends this task to step-by-step image sequences. Seer~\cite{gu2024seer} and AID~\cite{xing2025aid} predict videos from observed frames and language instructions. ShowMe~\cite{pu2026showme} models action--object state transformations, Demo2Tutorial~\cite{bai2026demo2tutorial} turns software recordings into multimodal tutorials, and PIE-V~\cite{loginova2026piev} generates mistakes and corrections in egocentric procedural videos. These works combine an observed scene with an instruction. \benchmark\ instead focuses on ordered laboratory procedures, visible state transitions, task-relevant contacts, safety constraints, and decision-support criteria tailored to procedural correctness.

\textbf{Video Generation Models (VGMs).} VGMs synthesize videos from text, images, or multimodal references, sometimes with explicit action or control signals. Open-weight models include Wan 2.2~\cite{wanvideo2025wan22}, LTX 2.3~\cite{ltx2026ltx2}, MiniMax~H3~\cite{minimax2026h3}, and HunyuanVideo 1.5~\cite{hunyuan2025hunyuanvideo15} for text-to-video (T2V) and image-to-video (I2V) generation. Cosmos3~\cite{nvidia2026cosmos3} and LingBot Video~\cite{ma2026lingbotvideo} extend this line to embodied and egocentric synthesis using multimodal, action, and robot data. Closed models such as Sora 2~\cite{openai2025sora2}, Gemini Omni Flash~\cite{googledeepmind2026omniflash}, Seedance 2.0~\cite{seedance2026seedance2}, and Kling AI 3.0~\cite{kuaishou2026kling3} add synchronized audio, multimodal conditioning, editing, and multi-shot control. Yet situated instruction requires more than realism: a video must execute the specified procedure and clearly expose its critical steps and visible state changes throughout the generated sequence.

\begin{figure}[t]
    \centering
    \includegraphics[width=1.\linewidth]{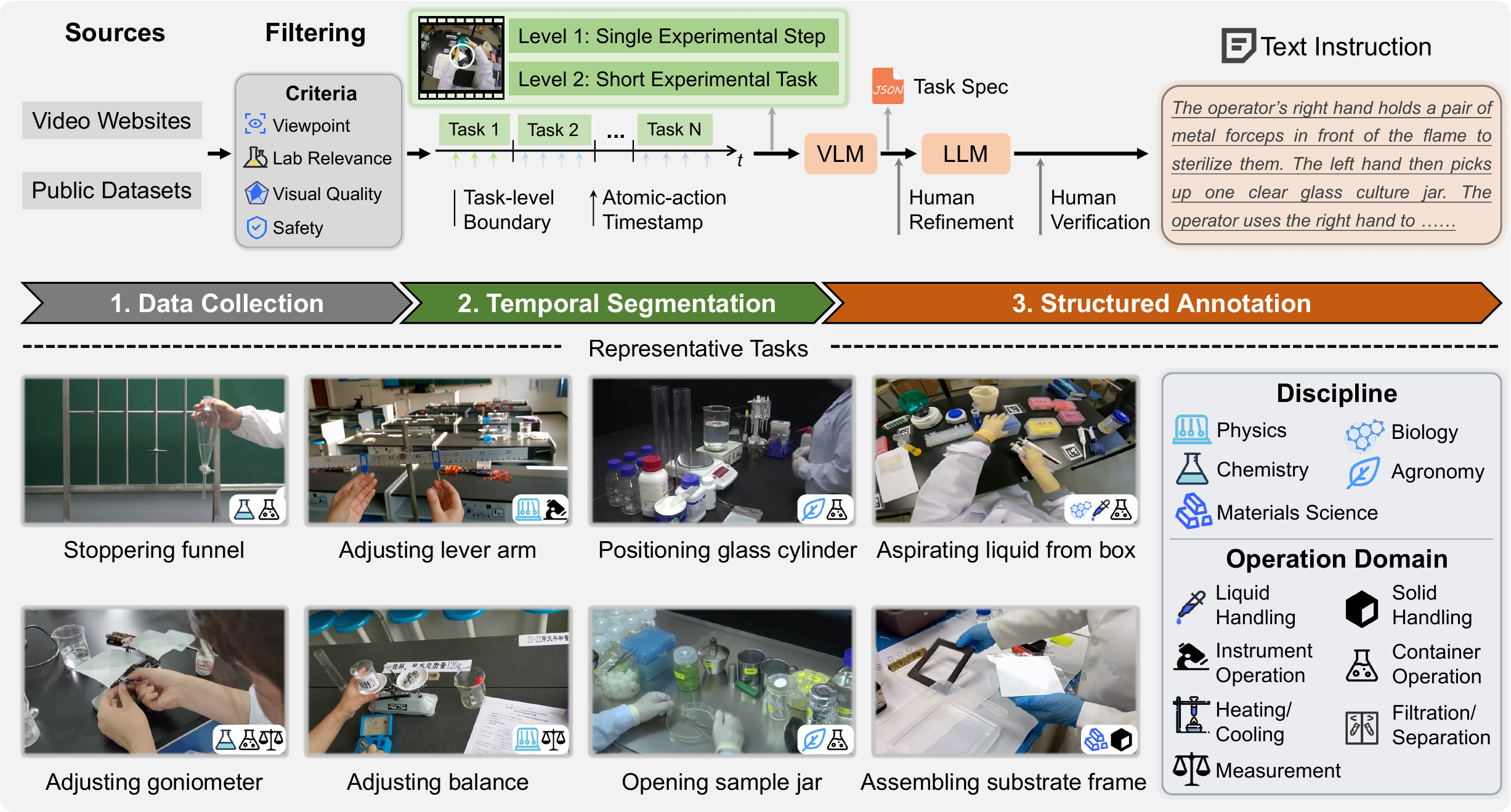}
    \caption{\textbf{\benchmark\ construction pipeline.} Laboratory videos are collected, filtered, segmented into tasks, and converted into human-verified specifications and instructions.}
    \label{fig:framework}
\end{figure}

\textbf{Scientific Video Generation Benchmarks.} Existing benchmarks assess synthesized physical or scientific outcomes. Morpheus~\cite{zhang2025morpheus} and Physics-IQ~\cite{motamed2026physicsiq} ground evaluation in controlled physical systems, VideoPhy-2~\cite{bansal2025videophy2} probes action-centric physical commonsense, and PhyGround~\cite{lin2026phyground} tests compliance with 13 physical laws. VideoScience-Bench~\cite{hu2025videoscience} evaluates undergraduate-level physical and chemical reasoning in T2V and I2V settings, while Sci-VBench~\cite{zhang2026scivbench} spans 1,253 expert-authored tasks across 60 scientific subjects. These benchmarks test whether a phenomenon unfolds correctly, not whether a model performs an ordered human procedure in a given workspace. \benchmark\ targets this gap through scene-conditioned human demonstrations paired with reference executions from the same workspace. Table~\ref{tab:benchmark_comparison} summarizes the distinction.

\section{The \benchmark\ Benchmark}

\subsection{Paradigm and Procedure Complexity}
\label{sec:taskdef}

For each task, we have the initial workspace observation $\textit{f}_0$, including relevant objects, spatial relationship, and scene context. Given $\textit{f}_0$ and a natural-language instruction $T$, the generator synthesizes $\textit{f}_{1:N}=\mathcal{G}(\textit{f}_0,T;\theta)$ while preserving workspace identity and continuity, executing the instructed actions in the correct order, and reaching the expected intermediate and final states. We organize tasks into two independently complexity strata: \textbf{Level~1 (Single Experimental Step)} contains 1--3 atomic actions centered on a single operation with a visually identifiable outcome and uses a nominal generation target of 4~s; \textbf{Level~2 (Short Experimental Task)} contains 4--10 atomic actions with at least one inter-step dependency, requires temporal ordering and state continuity across operations, and uses a nominal target of 12~s.
\looseness=-1

\subsection{Benchmark Construction}
\label{sec:dataset}

\paragraph{Data Sources and Reference Usage.}
Figure~\ref{fig:framework} summarizes the construction pipeline.
We select real laboratory procedure videos from Bilibili, YouTube, the FineBio dataset~\cite{yagi2025finebio}, and the ExpVid benchmark~\cite{xu2025expvid}. For each selected video, we record its URL or dataset identifier, source identifier, and scientific discipline, and retain this provenance information for all derived clips. The collected videos serve as reference videos and are treated as one valid realization of the underlying procedure rather than as a unique target trajectory. Therefore, evaluation will focus on required actions, causal order, observable state transitions, and task outcomes, while allowing alternative valid execution paths. Human verification will remove incidental reference behavior from the scoring criteria and records acceptable variants when necessary. Generated videos are also expected to preserve the identity and spatial geometry of the observed workspace, with unexplained viewpoint changes, cuts, motion discontinuities, or loss of the workspace treated as procedural failures. 

\paragraph{Filtering Criteria.}
Each candidate video is manually screened under four criteria. \textbf{Experiment Viewpoint} requires an operator-centered view that keeps the manipulation region and task-relevant objects visible. We retain first-person and nearby-observer views, annotate the viewpoint category, and report results separately for both. \textbf{Laboratory Relevance} requires a real, complete procedure with visually verifiable steps. \textbf{Visual Quality} requires clear, continuous footage with stable illumination and no persistent occlusion. \textbf{Safety Compliance} excludes procedures that depict unsafe operations or omit required protective measures. Full criteria are provided in Appendix~\ref{sec:filtering_criteria}. This filtering focuses \benchmark\ on short, observable, and verifiable procedures rather than attempting to represent all laboratory conditions, equipment configurations, or procedure types.

\begin{wrapfigure}{r}{0.50\linewidth}
    \centering
    \vspace{-10pt}
    \setlength{\belowcaptionskip}{-20pt}
    \includegraphics[width=\linewidth]{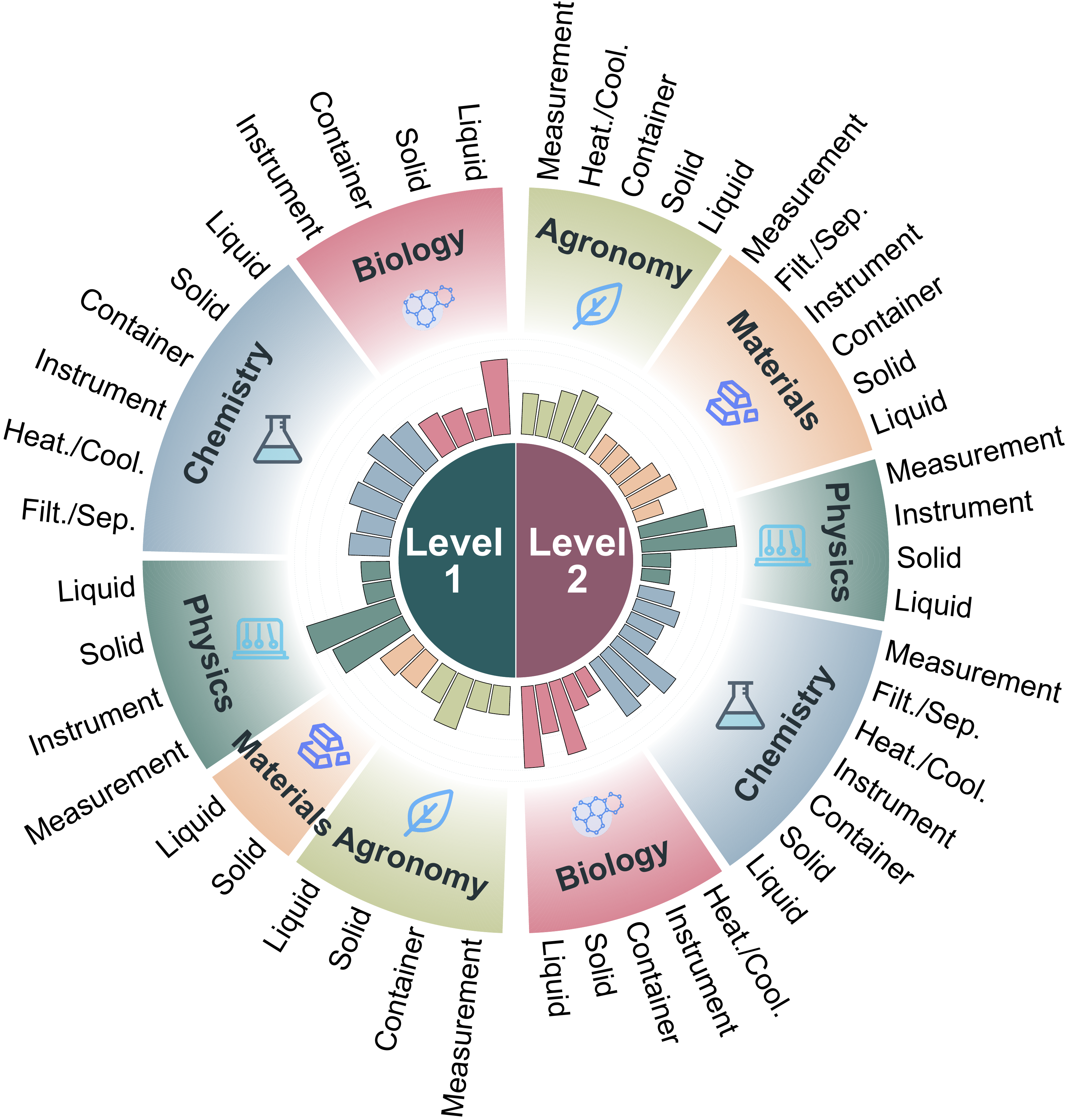}
    \caption{Distribution of the \numtasks{} benchmark tasks across task levels, scientific disciplines, and operation domains.}
    \label{fig:dataset_distribution}
\end{wrapfigure}

\paragraph{Data Statistics.}
Figure~\ref{fig:dataset_distribution} summarizes the benchmark composition. The benchmark comprises \numtasks{} tasks across \numdisciplines{} scientific disciplines, including \numlevelone{} Level~1 and \numleveltwo{} Level~2 clips. These tasks span seven operation domains, with instrument operations and liquid handling being the most prevalent. Because Level~2 tasks combine multiple experimental steps, a single task may involve more than one operation domain. Most clips are recorded from the nearby-observer viewpoint, with the remainder using the first-person viewpoint. Reference clips average 4.5 s for Level~1 and 12.7 s for Level~2. Additional dataset statistics and breakdowns are provided in Appendix~\ref{sec:dataset_statistics}.

\subsection{Task Annotation}
\label{sec:annotation}

\paragraph{Temporal Segmentation.}
We annotate each source video at two temporal resolutions aligned with the task levels defined in Sec.~\ref{sec:taskdef}. Level~1 intervals delimit individual experimental steps, while Level~2 intervals identify short experimental procedures that may consist of multiple consecutive steps. Every annotated interval records its precise start and end times. Level~1 and Level~2 annotations are defined independently, although a Level~2 interval may overlap with or contain one or more Level~1 intervals when the corresponding constituent steps are annotated. We use these temporal annotations to extract benchmark clips at the two task levels. Each clip begins with a stable and sufficiently clear view of the relevant workspace, and the key manipulations and resulting state transitions required by the task must remain clearly visible throughout the reference execution.

\paragraph{Structured Task Specification.}
For each clip, we construct a structured task specification using model-assisted annotation followed by human verification. Qwen3.6-27B~\cite{qwen2026qwen3} first generates a draft from the clip and a predefined template, which annotators then review and correct. The specification captures task-relevant objects and roles, spatial relations, ordered actions, interaction types, observable state transitions, and difficulty attributes. Actions are linked to stable object IDs and annotated with step index, acting hand when applicable, action verb, target object, contact type, and visible pre- and post-action states. Human verification ensures temporal alignment, object identity, action completeness and ordering, state-change accuracy, and consistency with the initial frame, while removing reference-specific behaviors unnecessary for successful execution.

\paragraph{Instruction Generation.}
After human verification, the task specification is frozen and used to generate the natural-language instruction provided to the video generator. Qwen3.6-27B converts the verified specification into a predefined instruction format. A human annotator then checks semantic consistency between the instruction and the specification. The final instruction preserves task-relevant object identities, action order, interactions, and observable state transitions, while abstracting away reference-specific details such as hand choice, exact motion trajectories, pauses, and other incidental execution behavior unless they are required for correctness. This design ensures that the instruction specifies what must be accomplished without constraining the generator to reproduce the particular trajectory shown in the reference video.

\begin{table}[t]
    \centering
    \caption{\textbf{Core Visual Quality} scores under GPT-5.6 Sol. \textbf{Obj.}, \textbf{Act.}, \textbf{St.}, and \textbf{Phys.} denote Object Consistency, Action Fidelity, State Correctness, and Physical Plausibility, respectively. Higher is better. Best values are bolded, and $S_{\mathrm{core}}$ is shaded by rank. $\dagger$ indicates commercial models.}
    \label{tab:vlm}
    \setlength{\tabcolsep}{9pt}
    \renewcommand{\arraystretch}{1}
    \resizebox{\textwidth}{!}{%
    \begin{tabular}{lcccccccccc}
        \toprule
        \multirow{2}{*}{\textbf{Model}} & \multicolumn{5}{c}{\textbf{Level~1}} & \multicolumn{5}{c}{\textbf{Level~2}} \\
        \arrayrulecolor{sepgray}\cmidrule(lr){2-6}\cmidrule(lr){7-11}\arrayrulecolor{black}
        & \textbf{Obj.} & \textbf{Act.} & \textbf{St.} & \textbf{Phys.} & $S_{\mathrm{core}}$
        & \textbf{Obj.} & \textbf{Act.} & \textbf{St.} & \textbf{Phys.} & $S_{\mathrm{core}}$ \\
        \midrule
        Wan 2.2
        & 93.7 & 59.4 & 79.0 & 90.1 & \hlcell{thirdfill}{76.2}
        & 89.8 & 37.3 & 58.2 & 84.3 & \hlcell{thirdfill}{60.0} \\

        LTX 2.3
        & 87.5 & 50.9 & 71.6 & 85.7 & \hlcell{thirdfill}{68.8}
        & 83.0 & 45.3 & 66.6 & 83.5 & \hlcell{thirdfill}{63.2} \\

        Cosmos3 Nano
        & 96.4 & 46.1 & 66.9 & 86.2 & \hlcell{thirdfill}{67.7}
        & 88.9 & 35.7 & 58.1 & 80.4 & \hlcell{thirdfill}{58.3} \\

        Cosmos3 Super
        & 94.8 & 51.7 & 71.3 & 86.0 & \hlcell{thirdfill}{71.0}
        & 90.8 & 37.1 & 57.5 & 83.3 & \hlcell{thirdfill}{59.5} \\

        LingBot Video
        & 92.7 & 48.0 & 67.8 & 85.8 & \hlcell{thirdfill}{68.2}
        & 88.7 & 39.9 & 61.7 & 83.6 & \hlcell{thirdfill}{61.7} \\

        MiniMax~H3
        & 96.7 & \textbf{69.2} & \textbf{84.6} & 93.1 & \hlcell{bestfill}{\textbf{82.6}}
        & 95.6 & 61.2 & 75.9 & 92.7 & \hlcell{thirdfill}{77.0} \\

        $\text{Seedance 2.0}^\dagger$
        & \textbf{98.5} & 58.4 & 79.9 & 92.7 & \hlcell{thirdfill}{77.9}
        & 94.6 & 67.2 & 77.9 & 93.2 & \hlcell{secondfill}{79.6} \\

        $\text{Wan 3.0}^\dagger$
        & 98.1 & 66.8 & 83.1 & \textbf{93.7} & \hlcell{secondfill}{81.6}
        & \textbf{95.7} & \textbf{67.4} & \textbf{80.3} & \textbf{93.7} & \hlcell{bestfill}{\textbf{80.3}} \\
        \bottomrule
    \end{tabular}}
\end{table}

\begin{table}[t]
    \centering
    \caption{\textbf{Completion} and \textbf{Overall} scores under GPT-5.6 Sol. \textbf{Comp.} denotes the completion factor. Higher is better. Best values are bolded, and Comp., Overall, and Pooled Overall are shaded by rank. \textbf{Pooled Overall} pools all tasks. $\dagger$ indicates commercial models.}
    \label{tab:summary}
    \setlength{\tabcolsep}{4pt}
    \renewcommand{\arraystretch}{1}
    \resizebox{\textwidth}{!}{%
    \begin{tabular}{lccccc ccccc c}
        \toprule
        \multirow{2}{*}{\textbf{Model}}
        & \multicolumn{5}{c}{\textbf{Level~1}}
        & \multicolumn{5}{c}{\textbf{Level~2}}
        & \multirow{2}{*}{\colhead{\textbf{Pooled}\\ \textbf{Overall}}}
        \\
        \arrayrulecolor{sepgray}\cmidrule(lr){2-6}\cmidrule(lr){7-11}\arrayrulecolor{black}
        & $S_{\mathrm{core}}$ & \textbf{Comp.} & $G_{\mathrm{scene}}$ & $G_{\mathrm{safety}}$ & \textbf{Overall}
        & $S_{\mathrm{core}}$ & \textbf{Comp.} & $G_{\mathrm{scene}}$ & $G_{\mathrm{safety}}$ & \textbf{Overall} &  \\
        \midrule
        Wan 2.2
        & 76.2 & \hlcell{thirdfill}{44.0} & 97.0 & 93.0 & \hlcell{thirdfill}{38.5}
        & 60.0 & \hlcell{thirdfill}{17.4} & 98.0 & 95.1 & \hlcell{thirdfill}{13.6}
        & \hlcell{thirdfill}{25.9} \\

        LTX 2.3
        & 68.8 & \hlcell{thirdfill}{27.2} & 96.0 & 87.1 & \hlcell{thirdfill}{22.1}
        & 63.2 & \hlcell{thirdfill}{25.2} & 93.2 & 89.3 & \hlcell{thirdfill}{22.1}
        & \hlcell{thirdfill}{22.1}  \\

        Cosmos3 Nano
        & 67.7 & \hlcell{thirdfill}{25.2} & 99.0 & 88.1 & \hlcell{thirdfill}{23.7}
        & 58.3 & \hlcell{thirdfill}{17.9} & 96.1 & 92.2 & \hlcell{thirdfill}{15.2}
        & \hlcell{thirdfill}{19.4}  \\

        Cosmos3 Super
        & 71.0 & \hlcell{thirdfill}{32.1} & 98.0 & 96.0 & \hlcell{thirdfill}{30.0}
        & 59.5 & \hlcell{thirdfill}{17.4} & 97.0 & 95.1 & \hlcell{thirdfill}{15.7}
        & \hlcell{thirdfill}{22.8}  \\

        LingBot Video
        & 68.2 & \hlcell{thirdfill}{23.2} & \textbf{100.0} & 92.0 & \hlcell{thirdfill}{20.0}
        & 61.7 & \hlcell{thirdfill}{22.3} & \textbf{99.0} & 95.1 & \hlcell{thirdfill}{19.1}
        & \hlcell{thirdfill}{19.5}  \\

        MiniMax~H3
        & \textbf{82.6} & \hlcell{bestfill}{\textbf{58.4}} & 98.0 & \textbf{98.0} & \hlcell{bestfill}{\textbf{53.6}}
        & 77.0 & \hlcell{thirdfill}{45.6} & 97.0 & 94.1 & \hlcell{thirdfill}{39.2}
        & \hlcell{bestfill}{\textbf{46.3}}  \\

        $\text{Seedance 2.0}^\dagger$
        & 77.9 & \hlcell{thirdfill}{45.5} & 98.0 & \textbf{98.0} & \hlcell{thirdfill}{41.9}
        & 79.6 & \hlcell{bestfill}{\textbf{53.3}} & 98.0 & \textbf{96.1} & \hlcell{bestfill}{\textbf{48.2}}
        & \hlcell{secondfill}{45.1} \\

        $\text{Wan 3.0}^\dagger$
        & 81.6 & \hlcell{secondfill}{50.9} & 98.0 & 93.0 & \hlcell{secondfill}{46.2}
        & \textbf{80.3} & \hlcell{secondfill}{52.9} & 88.3 & 94.1 & \hlcell{secondfill}{42.7}
        & \hlcell{thirdfill}{44.4}  \\
        \bottomrule
    \end{tabular}}
\end{table}

\subsection{Evaluation Protocol}
\label{sec:evaluation}
\paragraph{Evaluation Design.}
Our evaluation comprises three decoupled layers. \textbf{Core Visual Quality} includes \textit{Object Consistency}, \textit{Action Fidelity}, \textit{State Correctness}, and \textit{Physical Plausibility}, assessing object persistence, action correctness and order, state transitions, and physical interactions. \textbf{Critical Completion} measures whether essential procedural requirements are satisfied, while \textit{Scene Consistency} and \textit{Visual Safety} act as \textbf{Validity Constraints}; camera motion is allowed provided workspace continuity and safe execution are preserved. We use DeepSeek-v4-flash~\cite{deepseekai2026deepseekv4} to generate positively framed, visually verifiable checklist items, which are reviewed by human experts. This yields 3,229 verified question-answer items, averaging 13.0 and 18.6 per Level~1 and Level~2 task, respectively. Further details are provided in Appendix~\ref{sec:evaluation_design_details}.

\paragraph{Evaluation Criteria.}
Each checklist item is assigned one of three possible verdicts: \textsc{Yes}, \textsc{No}, or \textsc{Unjudgeable}. Invalid, ambiguous, or inapplicable items are removed during expert verification. For the four core dimensions, both \textsc{No} and \textsc{Unjudgeable} are treated as negative evidence. Their importance-weighted violations are aggregated within each dimension and mapped to a dimension-level score using the penalty function defined in Appendix~\ref{sec:scoring_procedure}. The resulting four dimension scores are then combined using their geometric mean to obtain $S_{\mathrm{core}}$. Task completion is assessed exclusively from \emph{Critical} core items. Let $n_c$ denote the number of \emph{Critical} items that are not judged \textsc{Yes}. We define $C(n_c)=1$, $0.5$, and $0$ for $n_c=0$, $1$, and $n_c\geq2$, respectively. Scene consistency and visual safety act as binary validity gates: either gate is closed by an explicit \textsc{No}, whereas \textsc{Unjudgeable} alone does not trigger a veto. The final score is therefore
\begin{equation}
S_{\mathrm{overall}} = S_{\mathrm{core}} \cdot C(n_c) \cdot G_{\mathrm{scene}} \cdot G_{\mathrm{safety}}.
\label{eq:overall}
\end{equation}
Scores are computed for each video and averaged within each task level. Scene consistency and visual safety are additionally reported as separate gate validity rates.

\begin{table}[t]
    \centering
    \caption{\textbf{Core Visual Quality} scores under human evaluation. Higher is better. Best values are bolded, and $S_{\mathrm{core}}$ is shaded by rank. $\dagger$ indicates commercial models.}
    \label{tab:vlm_human}
    \setlength{\tabcolsep}{9pt}
    \renewcommand{\arraystretch}{1}
    \resizebox{\textwidth}{!}{%
    \begin{tabular}{lcccccccccc}
        \toprule
        \multirow{2}{*}{\textbf{Model}} & \multicolumn{5}{c}{\textbf{Level~1}} & \multicolumn{5}{c}{\textbf{Level~2}} \\
        \arrayrulecolor{sepgray}\cmidrule(lr){2-6}\cmidrule(lr){7-11}\arrayrulecolor{black}
        & \textbf{Obj.} & \textbf{Act.} & \textbf{St.} & \textbf{Phys.} & $S_{\mathrm{core}}$
        & \textbf{Obj.} & \textbf{Act.} & \textbf{St.} & \textbf{Phys.} & $S_{\mathrm{core}}$ \\
        \midrule
        Wan 2.2
        & 86.2 & 53.4 & 66.6 & 87.1 & \hlcell{thirdfill}{68.4}
        & 77.0 & 30.2 & 48.9 & 79.3 & \hlcell{thirdfill}{51.4} \\

        LTX 2.3
        & 56.6 & 43.6 & 57.2 & 73.5 & \hlcell{thirdfill}{53.5}
        & 40.9 & 24.1 & 47.7 & 70.4 & \hlcell{thirdfill}{39.7} \\

        Cosmos3 Nano
        & 86.0 & 40.7 & 56.0 & 77.8 & \hlcell{thirdfill}{59.3}
        & 61.5 & 20.4 & 40.7 & 69.8 & \hlcell{thirdfill}{40.9} \\

        Cosmos3 Super
        & 89.1 & 50.1 & 61.9 & 85.2 & \hlcell{thirdfill}{66.0}
        & 76.2 & 24.3 & 46.6 & 74.2 & \hlcell{thirdfill}{47.6} \\

        LingBot Video
        & 81.6 & 45.2 & 61.0 & 79.0 & \hlcell{thirdfill}{61.3}
        & 53.9 & 23.2 & 47.6 & 68.9 & \hlcell{thirdfill}{42.1} \\

        MiniMax~H3
        & \textbf{95.5} & \textbf{70.9} & \textbf{80.2} & \textbf{93.5} & \hlcell{bestfill}{\textbf{81.6}}
        & \textbf{94.6} & \textbf{63.8} & \textbf{74.6} & \textbf{93.2} & \hlcell{bestfill}{\textbf{77.3}} \\

        $\text{Seedance 2.0}^\dagger$
        & 91.5 & 70.5 & 77.2 & 90.5 & \hlcell{secondfill}{79.1}
        & 90.1 & 60.5 & 69.3 & \textbf{93.2} & \hlcell{thirdfill}{73.9} \\

        $\text{Wan 3.0}^\dagger$
        & 89.8 & 66.1 & 73.5 & 89.8 & \hlcell{thirdfill}{75.7}
        & 87.9 & 62.7 & 74.1 & 92.2 & \hlcell{secondfill}{74.6} \\
        \bottomrule
    \end{tabular}}
\end{table}

\begin{table}[t]
    \centering
    \caption{\textbf{Completion} and \textbf{Overall} scores under human evaluation. Higher is better. Best values are bolded, and Comp., Overall, and Pooled Overall are shaded by rank. $\dagger$ indicates commercial models.}
    \label{tab:summary_human}
    \setlength{\tabcolsep}{4pt}
    \renewcommand{\arraystretch}{1}
    \resizebox{\textwidth}{!}{%
    \begin{tabular}{lccccc ccccc c}
        \toprule
        \multirow{2}{*}{\textbf{Model}}
        & \multicolumn{5}{c}{\textbf{Level~1}}
        & \multicolumn{5}{c}{\textbf{Level~2}}
        & \multirow{2}{*}{\colhead{\textbf{Pooled}\\ \textbf{Overall}}} \\
        \arrayrulecolor{sepgray}\cmidrule(lr){2-6}\cmidrule(lr){7-11}\arrayrulecolor{black}
        & $S_{\mathrm{core}}$ & \textbf{Comp.} & $G_{\mathrm{scene}}$ & $G_{\mathrm{safety}}$ & \textbf{Overall}
        & $S_{\mathrm{core}}$ & \textbf{Comp.} & $G_{\mathrm{scene}}$ & $G_{\mathrm{safety}}$ & \textbf{Overall} & \\
        \midrule
        Wan 2.2
        & 68.4 & \hlcell{thirdfill}{27.2} & 99.0 & 84.1 & \hlcell{thirdfill}{23.3}
        & 51.4 & \hlcell{thirdfill}{8.20} & 99.0 & 75.7 & \hlcell{thirdfill}{7.70}
        & \hlcell{thirdfill}{15.4} \\

        LTX 2.3
        & 53.5 & \hlcell{thirdfill}{7.40} & 93.0 & 73.2 & \hlcell{thirdfill}{5.30}
        & 39.7 & \hlcell{thirdfill}{0.90} & 91.2 & 72.8 & \hlcell{thirdfill}{0.70}
        & \hlcell{thirdfill}{3.00} \\

        Cosmos3 Nano
        & 59.3 & \hlcell{thirdfill}{12.8} & 99.0 & 79.2 & \hlcell{thirdfill}{10.4}
        & 40.9 & \hlcell{thirdfill}{1.40} & 98.0 & 69.9 & \hlcell{thirdfill}{1.30}
        & \hlcell{thirdfill}{5.80} \\

        Cosmos3 Super
        & 66.0 & \hlcell{thirdfill}{21.7} & 99.0 & 85.1 & \hlcell{thirdfill}{18.9}
        & 47.6 & \hlcell{thirdfill}{5.30} & 97.0 & 76.6 & \hlcell{thirdfill}{4.20}
        & \hlcell{thirdfill}{11.5} \\

        LingBot Video
        & 61.3 & \hlcell{thirdfill}{19.3} & 96.0 & 81.1 & \hlcell{thirdfill}{18.0}
        & 42.1 & \hlcell{thirdfill}{3.80} & 92.2 & 72.8 & \hlcell{thirdfill}{3.20}
        & \hlcell{thirdfill}{10.5} \\

        MiniMax~H3
        & \textbf{81.6} & \hlcell{bestfill}{\textbf{51.9}} & \textbf{100.0} & \textbf{98.0} & \hlcell{bestfill}{\textbf{49.7}}
        & \textbf{77.3} & \hlcell{bestfill}{\textbf{43.6}} & 99.0 & 92.2 & \hlcell{bestfill}{\textbf{39.0}}
        & \hlcell{bestfill}{\textbf{44.3}} \\

        $\text{Seedance 2.0}^\dagger$
        & 79.1 & \hlcell{secondfill}{50.9} & \textbf{100.0} & 97.0 & \hlcell{secondfill}{48.5}
        & 73.9 & \hlcell{thirdfill}{39.3} & \textbf{100.0} & \textbf{96.1} & \hlcell{secondfill}{37.7}
        & \hlcell{secondfill}{43.0} \\

        $\text{Wan 3.0}^\dagger$
        & 75.7 & \hlcell{thirdfill}{42.0} & 98.0 & 91.0 & \hlcell{thirdfill}{38.8}
        & 74.6 & \hlcell{secondfill}{39.8} & 96.1 & 95.1 & \hlcell{thirdfill}{35.8}
        & \hlcell{thirdfill}{37.3} \\
        \bottomrule
    \end{tabular}}
\end{table}

\section{Experiments}

\subsection{Experimental Setup}
\label{sec:experimental_setup}
We evaluate \nummodels{} models, including 6 open-weight models: \textbf{Wan 2.2 I2V A14B}~\cite{wanvideo2025wan22} (simplified as Wan 2.2), \textbf{LTX 2.3}~\cite{ltx2026ltx2}, \textbf{Cosmos3 Nano}, \textbf{Cosmos3 Super}~\cite{nvidia2026cosmos3}, \textbf{LingBot Video}~\cite{ma2026lingbotvideo}, and \textbf{MiniMax~H3}~\cite{minimax2026h3}, as well as 2 commercial models: \textbf{Seedance 2.0}~\cite{seedance2026seedance2} and \textbf{Wan 3.0}~\cite{alibaba2026wan30}. All models are provided with the same initial frame and a verified natural-language instruction to ensure a controlled comparison across models for each task. To prioritize broad coverage across both tasks and generators, we evaluate one generated sample per model--task pair under fixed model-specific generation settings. The exact model checkpoints, release dates, and generation configurations used in our experiments are reported in Appendix~\ref{sec:model_details}. For automated evaluation, we provide GPT-5.6~Sol with the generated video, sampled at 4 frames per second (FPS), together with the corresponding evaluation rules and task-specific checklist. The evaluator is then instructed to produce a categorical verdict for each checklist item.

\begin{figure}[t]
    \centering
    \begin{minipage}[c]{0.48\linewidth}
        \centering
        \includegraphics[width=\linewidth]{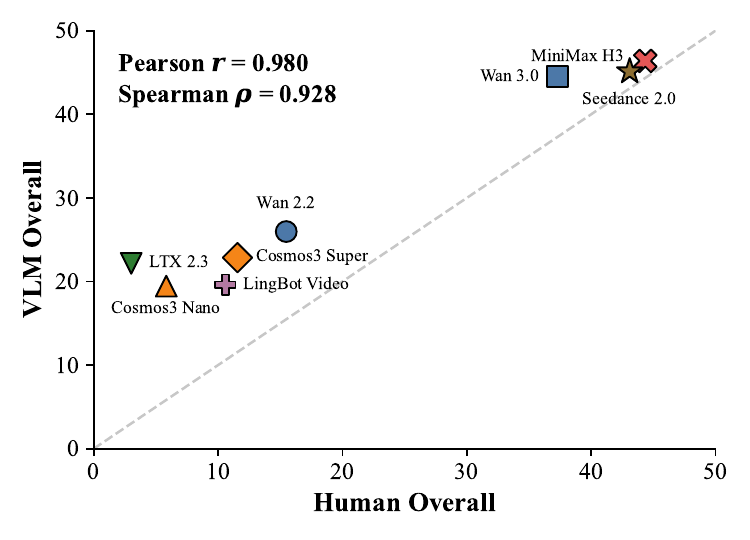}
        {
        \setlength{\abovecaptionskip}{-15pt}
        \caption{Overall score consistancy of each model under Human and VLM evaluation.}
        \label{fig:human_vlm_agreement}}
    \end{minipage}\hfill
    \begin{minipage}[c]{0.52\linewidth}
        \centering
        \captionof{table}{Agreement between human and VLM scores for all \nummodels{} generators. Metrics are correlation, mean absolute error (MAE), maximum rank shift, and top-three overlap. \textbf{Overall} pools all \numtasks{} tasks.}
        \label{tab:human_vlm_agreement}
        {\footnotesize
        \setlength{\belowcaptionskip}{1pt}
        \setlength{\tabcolsep}{4pt}
        \renewcommand{\arraystretch}{1.1}
        \begin{tabular}{lccc}
            \toprule
            \textbf{Score} & \textbf{Level~1} & \textbf{Level~2} & \textbf{Overall} \\
            \midrule
            Pearson $r$ & 0.903 & 0.942 & 0.980 \\
            Spearman $\rho$ & 0.904 & 0.523 & 0.928 \\
            MAE & 9.54 & 10.77 & 9.35 \\
            Max $\Delta$Rank & 2 & 4 & 2 \\
            Top-3 Overlap & 3/3 & 3/3 & 3/3 \\
            \bottomrule
        \end{tabular}}
    \end{minipage}
\end{figure}

\begin{figure}[t]
    \centering
    \includegraphics[width=0.98\linewidth]{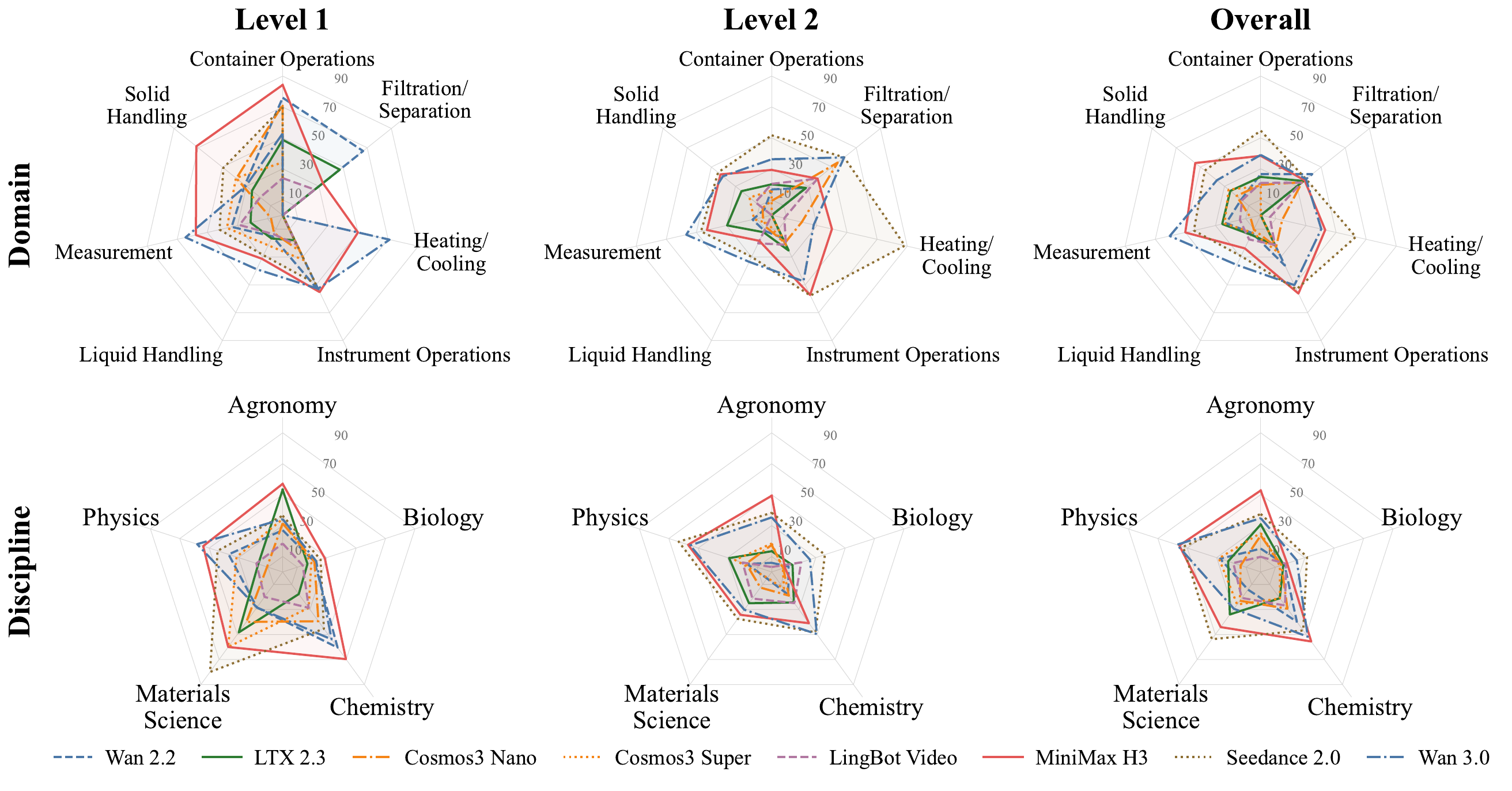}
    \caption{VLM evaluation scores broken down by operation domain (top row) and scientific discipline (bottom row) for Level~1, Level~2, and Overall.}
    \label{fig:radar_overall}
\end{figure}

\subsection{Experimental Results}
\label{sec:main_results}

\paragraph{GPT-5.6 Sol Results.}
Tables~\ref{tab:vlm} and~\ref{tab:summary} show that procedural execution is the main bottleneck across models. Object consistency and physical plausibility remain relatively strong, whereas action fidelity, state correctness, and especially completion degrade on Level~2. The drop is particularly pronounced for weaker systems, whose videos often remain visually coherent despite failing to execute all required steps or realize the intended final state. As a result, high scene and safety validity does not necessarily translate into high Overall performance. MiniMax~H3, Seedance~2.0, and Wan~3.0 form the strongest group overall, although their relative ordering varies across complexity levels and metrics. Notably, MiniMax~H3 leads on Level~1, while Wan~3.0 achieves the highest Level~2 core score and Seedance~2.0 the highest Level~2 Overall score. These results indicate that current models preserve appearance and local physical plausibility more reliably than they execute temporally ordered procedures and achieve the required state transitions.

\paragraph{Human Evaluation Results.}
Human evaluation in Tables~\ref{tab:vlm_human} and~\ref{tab:summary_human} broadly confirms the same conclusions, while changing some fine-grained rankings. MiniMax~H3, Seedance~2.0, and Wan~3.0 remain the leading systems, and Level~2 again exposes the largest weaknesses in action execution and procedural completion. Compared with human judgments, GPT-5.6~Sol is generally more permissive, particularly for weaker models, assigning substantially higher completion and Overall scores; for example, pooled Overall drops from 22.1 to 3.0 for LTX~2.3 and from 19.4 to 5.8 for Cosmos3 Nano under human evaluation. The agreement in overall trends suggests that automated evaluation captures the main performance structure, while human judgments apply a stricter standard for whether the requested procedure is actually completed.

\subsection{Further Analysis}
\label{sec:further_analysis}

\paragraph{Human--VLM Agreement.}
Human and VLM evaluations show strong generator-level agreement (Figure~\ref{fig:human_vlm_agreement}; Table~\ref{tab:human_vlm_agreement}). Pearson correlation is high at both Level~1 ($r=0.903$) and Level~2 ($r=0.942$), and increases to $r=0.980$ after pooling, which also yields the lowest MAE (9.35). Rank agreement is weaker on Level~2 ($\rho=0.523$, maximum rank shift $=4$) but becomes substantially more stable after aggregation ($\rho=0.928$, maximum rank shift $=2$). Notably, the top-3 set is preserved across all settings, and the pooled human and VLM rankings agree on the ordering of the three strongest generators. Figure~\ref{fig:human_vlm_agreement} further shows that larger deviations occur mainly among lower-scoring models. Overall, aggregation provides stable human--VLM agreement for generator comparison, particularly among the strongest models, with additional agreement analyses reported in Appendix~\ref{sec:human_evaluation_details}.

\begin{table}[t]
\centering
\small
\renewcommand{\arraystretch}{1.05}
\caption{Cross-evaluator consensus tiers across task levels and evaluators. Subscripts indicate ranks.}
\label{tab:consensus_tiers}
\resizebox{\textwidth}{!}{
\begin{tabular}{lcccccccc}
\toprule
\multirow{2}{*}{\textbf{Model}} & \multicolumn{3}{c}{\textbf{Human Evaluation}} & \multicolumn{3}{c}{\textbf{VLM Evaluation}} & \multicolumn{2}{c}{\textbf{Consensus}} \\
\cmidrule(lr){2-4}\cmidrule(lr){5-7}\cmidrule(lr){8-9}
& Level 1 & Level 2 & Overall & Level 1 & Level 2 & Overall & Mean Rank & Tier \\
\midrule
MiniMax~H3 & 49.7$_1$ & 39.0$_1$ & 44.3$_1$ & 53.6$_1$ & 39.2$_3$ & 46.3$_1$ & 1.33 & \cellcolor{bestfill}\textbf{T0} \\
Seedance~2.0$^\dagger$ & 48.5$_2$ & 37.7$_2$ & 43.0$_2$ & 41.9$_3$ & 48.2$_1$ & 45.1$_2$ & 2.00 & \cellcolor{bestfill}\textbf{T0} \\
Wan~3.0$^\dagger$ & 38.8$_3$ & 35.8$_3$ & 37.3$_3$ & 46.2$_2$ & 42.7$_2$ & 44.4$_3$ & 2.67 & \cellcolor{bestfill}\textbf{T0} \\
\midrule
Wan~2.2 & 23.3$_4$ & 7.70$_4$ & 15.4$_4$ & 38.5$_4$ & 13.6$_8$ & 25.9$_4$ & 4.67 & \cellcolor{secondfill}T1 \\
Cosmos3~Super & 18.9$_5$ & 4.20$_5$ & 11.5$_5$ & 30.0$_5$ & 15.7$_6$ & 22.8$_5$ & 5.17 & \cellcolor{secondfill}T1 \\
\midrule
LingBot~Video & 18.0$_6$ & 3.20$_6$ & 10.5$_6$ & 20.0$_8$ & 19.1$_5$ & 19.5$_7$ & 6.33 & \cellcolor{thirdfill}T2 \\
LTX~2.3 & 5.30$_8$ & 0.70$_8$ & 3.00$_8$ & 22.1$_7$ & 22.1$_4$ & 22.1$_6$ & 6.83 & \cellcolor{thirdfill}T2 \\
Cosmos3~Nano & 10.4$_7$ & 1.30$_7$ & 5.80$_7$ & 23.7$_6$ & 15.2$_7$ & 19.4$_8$ & 7.00 & \cellcolor{thirdfill}T2 \\
\bottomrule
\end{tabular}
}
\end{table}

\paragraph{Domain and Discipline Breakdown.}
Figure~\ref{fig:radar_overall} shows substantial variation across operation domains and scientific disciplines. MiniMax~H3 and Wan~3.0 provide the broadest coverage, with complementary strengths: MiniMax~H3 performs strongly in instrument operations, solid handling, agronomy, and chemistry, while Wan~3.0 leads in liquid handling, measurement, and physics. Seedance~2.0 instead excels in container operations, heating/cooling, and materials science, and Wan~2.2 is strongest in filtration/separation. Across models, Level~2 scores generally decline across multiple domains and disciplines rather than in a single category, showing that the increased difficulty is distributed across the benchmark rather than concentrated in a specific operation type.

\paragraph{Failure Attribution.}
We inspect representative failures and identify several recurring patterns across models.
\textbf{1) Procedural Omission and Misordering.}
The most common failures occur when models omit required operations, execute them out of order, or apply them to the wrong object. These errors often arise despite visually coherent generation, highlighting procedural execution as a more persistent bottleneck than appearance preservation.
\textbf{2) Incomplete State Realization.}
Models may perform an action without producing the required observable consequence, such as a liquid-level change, instrument reading, or final object configuration. Such errors may follow earlier action failures or reflect difficulty in realizing fine-grained quantitative or positional outcomes.
\textbf{3) Object Drift.}
Task-relevant objects may deform, disappear, merge, or become difficult to track, while occasional scene discontinuities further disrupt procedural continuity. These failures are less common than action errors but can still invalidate otherwise plausible executions.

\paragraph{Cross-Evaluator Consensus Tiers.}
The re-sampling analysis in Appendix~\ref{sec:resampling} shows that the exact ordering among the strongest models may vary with generation randomness. We therefore summarize performance using a rank-based consensus analysis across six evaluation views: Level~1, Level~2, and Overall scores from both human and VLM evaluators. Models are ranked in each view, averaged by rank, and grouped into three tiers using the two largest adjacent gaps. As shown in Table~\ref{tab:consensus_tiers}, MiniMax~H3, Seedance~2.0, and Wan~3.0 consistently form the top tier (T0), remaining within the top three across all six views, while Wan~2.2 and Cosmos3 Super form T1 and the remaining models form T2. Together with the re-sampling results, this indicates that the separation between performance tiers is more stable than the exact ordering within T0.


\section{Conclusion}
We introduce \benchmark, a benchmark for evaluating procedural correctness in laboratory instructional video generation. Experiments on \nummodels\ frontier models reveal a clear gap between visual fidelity and procedural success: models often preserve scenes and objects but miss, reorder, or obscure required actions and state changes. Completion is the main bottleneck on Level~2 tasks, whose longer procedures require sustained temporal ordering and state continuity. Overall, current video generation models remain unreliable as standalone sources of laboratory instruction, and \benchmark\ provides a systematic framework for evaluating and diagnosing procedural failures.

\bibliographystyle{unsrtnat}
\bibliography{main}

\clearpage
\appendix

\startcontents[appendices]
\printcontents[appendices]{l}{1}{
  \section*{Appendix Table of Contents}
  \setcounter{tocdepth}{2}
}
\clearpage

\section{Benchmark Construction Details}

\subsection{Source Video Filtering Criteria}
\label{sec:filtering_criteria}

We manually screen every candidate source video against the following four eligibility criteria:
\begin{itemize}
    \item \textbf{Experiment Viewpoint.} The video must provide an operator-centered view in which the manipulation region and task-relevant objects remain clearly visible. We retain first-person views recorded near the operator's eye level and nearby-observer views captured close to the operator and approximately aligned with the workspace.
    \item \textbf{Laboratory Relevance.} The video must depict a real laboratory procedure rather than an animation, non-experimental demonstration, or everyday activity. Experimental steps and their outcomes must be clearly discernible and visually verifiable.
    \item \textbf{Visual Quality.} The footage must be clear and continuous, with stable illumination and no severe camera shake or persistent occlusion. Videos with cuts, camera switches, ambiguous state changes, substantial editing, excessive text or branding, or severely cluttered backgrounds are excluded. We prefer videos in which all task-relevant apparatus is already visible in the initial workspace frame used for model conditioning.
    \item \textbf{Safety Compliance.} The video must not depict dangerous or clearly unsafe operations, omit required protective measures, or otherwise encourage unsafe practice.
\end{itemize}

\begin{table}[ht]
    \centering
    \caption{Composition of the \benchmark\ benchmark by video source.}
    \setlength{\tabcolsep}{15pt}
    \label{tab:video_sources}
    \begin{tabular}{lcc}
        \toprule
        \textbf{Source} & \textbf{\# Videos} & \textbf{\# Task Clips} \\
        \midrule
        Bilibili & 61 & 163 \\
        FineBio~\citep{yagi2025finebio} & 15 & 23 \\
        ExpVid~\citep{xu2025expvid} & 3 & 14 \\
        YouTube & 2 & 4 \\
        \midrule
        \textbf{Total} & \textbf{81} & \textbf{\numtasks} \\
        \bottomrule
    \end{tabular}
\end{table}

\subsection{Source Complementarity}
The \benchmark\ benchmark draws from 4 complementary video sources: Bilibili, YouTube, FineBio~\cite{yagi2025finebio}, and ExpVid~\cite{xu2025expvid}. These sources provide complementary coverage in both scientific content and recording viewpoint. FineBio contributes primarily first-person biology laboratory procedures and constitutes the main source of our biology tasks. In contrast, Bilibili and YouTube broaden the disciplinary coverage beyond biology, particularly for physics- and chemistry-oriented experiments, while also providing a diverse range of laboratory operations and workspace configurations. ExpVid, which is derived from professionally recorded JoVE experimental demonstrations, further complements the web-sourced videos with structured, multi-step laboratory procedures. We therefore curate across sources rather than relying on any single dataset or platform. After filtering and temporal annotation, the 81 retained source videos yield \numtasks{} benchmark task clips. Table~\ref{tab:video_sources} summarizes the contribution of each source.

\begin{figure}[ht]
    \centering
    \includegraphics[width=1.0\linewidth]{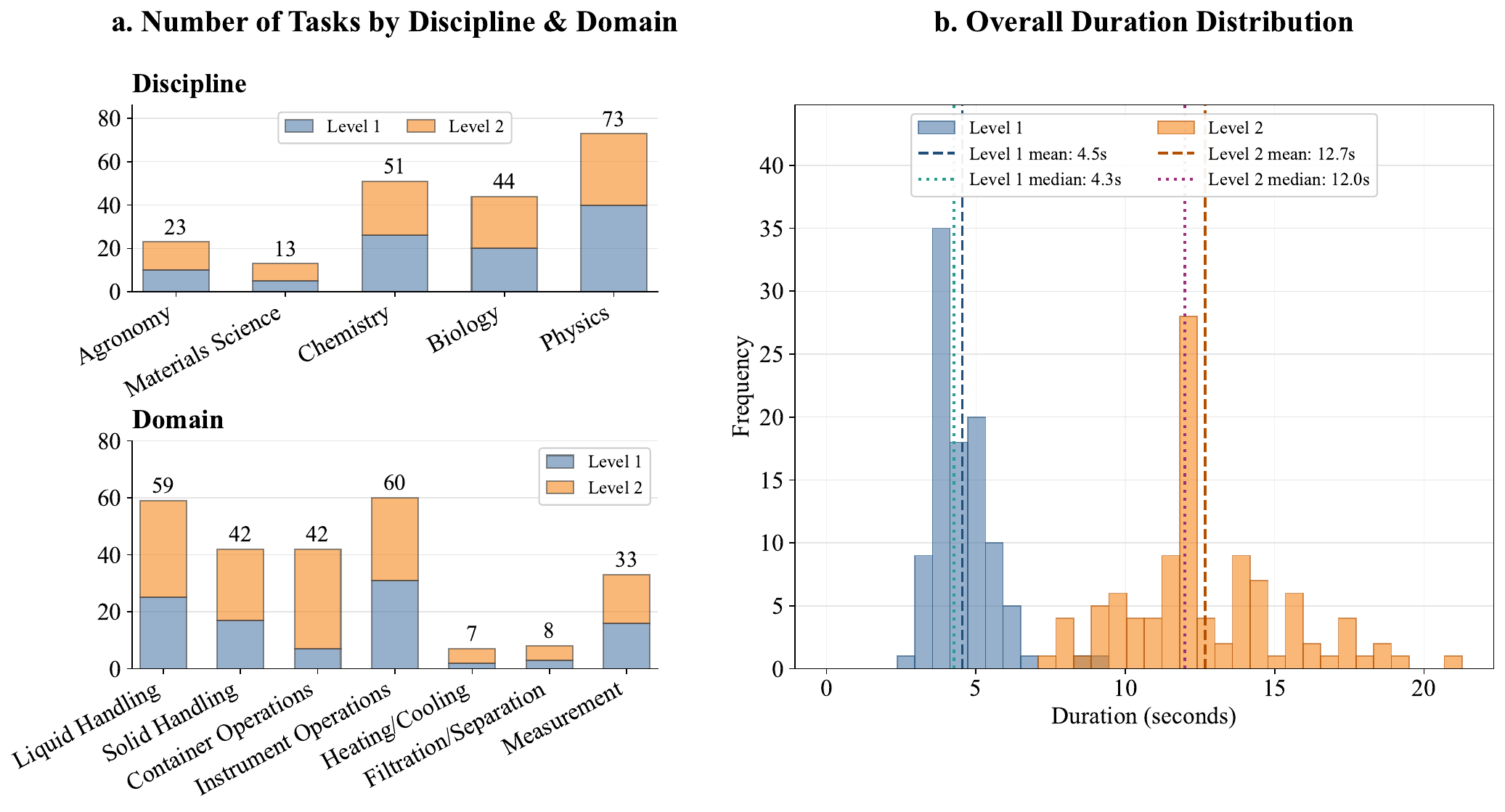}
    \caption{Benchmark task composition by discipline and domain, and reference-clip duration distributions across Level~1 and Level~2 tasks.}
    \label{fig:dataset_statistics}
\end{figure}

\subsection{Dataset Statistics}
\label{sec:dataset_statistics}

Figure~\ref{fig:dataset_statistics} provides a detailed breakdown of the benchmark composition and reference-clip durations. Across scientific disciplines, the benchmark contains 73 tasks in Physics, 51 in Chemistry, 44 in Biology, 23 in Agronomy, and 13 in Materials Science. In terms of operation domains, it includes 60 tasks involving Instrument Operations, 59 involving Liquid Handling, 42 involving Solid Handling, 42 involving Container Operations, 33 involving Measurement, 8 involving Filtration/Separation, and 7 involving Heating/Cooling. These domains cover a range of laboratory operations from routine manipulation of instruments, liquids, solids, and containers to measurement, thermal processing, and separation procedures. Since a task, particularly at Level~2, may involve multiple types of experimental operations, we assign operation domains in a multi-label manner, and the corresponding counts therefore do not sum to the total number of tasks.

\begin{figure}[t]
    \centering
    \includegraphics[width=0.92\linewidth]{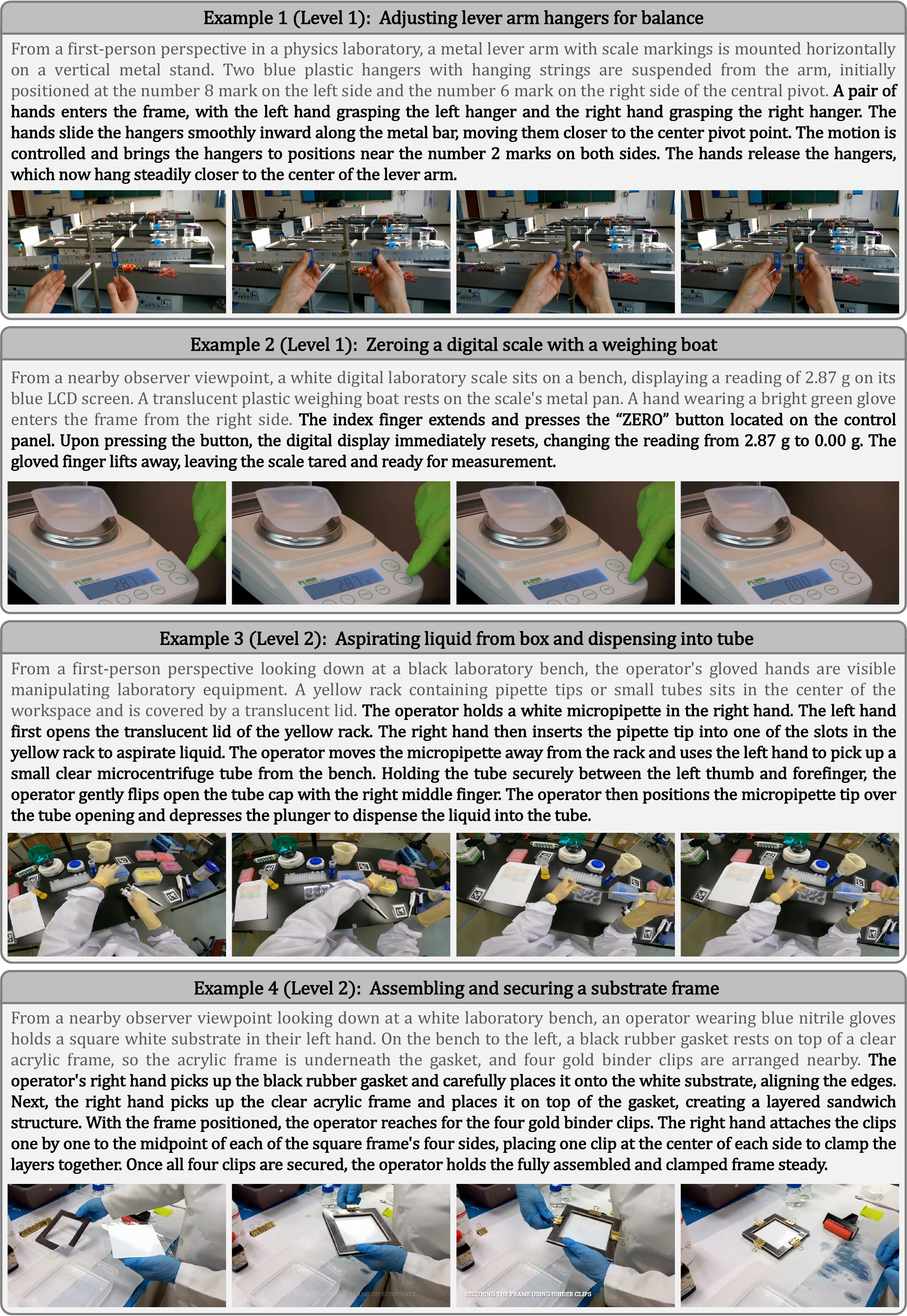}
    \caption{Four example tasks from \benchmark. Each block shows the task title, the difficulty level, the rendered generation instruction, and four frames sampled from the reference clip.}
    \label{fig:task_examples}
\end{figure}

Reference-clip durations are also organized differently across the two task levels. Level~1 clips are concentrated around 3--6 s, with a mean duration of 4.5 s and a median of 4.3 s. Level~2 clips span a broader range, typically around 8--16 s with a longer tail, with a mean duration of 12.7 s and a median of 12.0 s. In our benchmark design, Level~1 tasks correspond to localized experimental steps, while Level~2 tasks cover longer procedures that contain multiple temporally connected operations. The resulting difference in clip duration follows directly from this task-level definition. Figure~\ref{fig:task_examples} presents four completed benchmark instances from the two levels.

\clearpage
\clearpage

\section{Evaluation Details}

\subsection{Evaluation Design Details}
\label{sec:evaluation_design_details}

This section provides additional details on the evaluation protocol summarized in Sec.~\ref{sec:evaluation}. 

\paragraph{Dimension Definitions.}
Each task is associated with a task-specific checklist covering six dimensions: \textit{Object Consistency}, \textit{Action Fidelity}, \textit{State Correctness}, \textit{Physical Plausibility}, \textit{Scene Consistency}, and \textit{Visual Safety}. The first four dimensions measure the \textbf{Core Visual Quality} of generated videos, while the latter two act as \textbf{Validity Constraints}. \textbf{Critical Completion} is derived from the critical items within the four core dimensions rather than treated as an independent dimension.
\textit{Object Consistency} evaluates whether task-relevant objects preserve their identity and appearance throughout the video. \textit{Action Fidelity} assesses whether required actions are visibly executed on the correct targets and in the required order. \textit{State Correctness} measures whether the expected observable state transitions and final states are achieved. \textit{Physical Plausibility} evaluates whether object interactions and material behavior remain physically credible. \textit{Scene Consistency} checks the temporal continuity of the workspace, viewpoint, and visual environment, while allowing camera motion consistent with the recorded viewpoint. \textit{Visual Safety} evaluates task-relevant, visually observable safety conditions, such as containment, apparatus integrity, and safe object handling.

\paragraph{Checklist Construction.}
We generate one checklist per task using DeepSeek-v4-flash~\cite{deepseekai2026deepseekv4} with a fixed prompt template and JSON-constrained decoding (Appendix~\ref{sec:qa_generation_prompt}). The model receives only the task instruction, identifier, and difficulty level; reference videos and generated outputs are not provided. Checklist size is adapted to task complexity. Each item is required to express a single positively phrased and visually verifiable claim grounded in the task instruction. Required actions, observable state transitions, ordering constraints, object consistency, physical interactions, and relevant safety conditions are explicitly covered. Items are additionally assigned an importance level of \emph{Critical}, \emph{Standard}, or \emph{Supplementary}.

\paragraph{Expert Verification.}
Generated checklists are first screened using deterministic validation rules for schema validity, dimensional coverage, duplication, and required importance labels. They are then reviewed by domain experts against the task instruction and reference video. Reviewers verify visual decidability, task relevance, non-redundancy, and execution neutrality, and revise or remove items when necessary. The verified checklist for each task is subsequently frozen and shared across automated evaluation, human evaluation, and sensitivity analyses.

\paragraph{Checklist Statistics.}
The resulting benchmark contains 3,229 verified checklist items across \numtasks{} tasks, including 1,312 items for \numlevelone{} Level~1 tasks and 1,917 items for \numleveltwo{} Level~2 tasks. This corresponds to an average of 13.0 items per Level~1 task and 18.6 items per Level~2 task. All tasks are covered by all six evaluation dimensions. Action Fidelity constitutes the largest fraction of the checklist corpus, reflecting the greater number of atomic actions and inter-step ordering dependencies required by multi-step procedures with multiple dependent actions.

\subsection{Formal Scoring Procedure}
\label{sec:scoring_procedure}

This appendix formalizes the scoring procedure summarized in Sec.~\ref{sec:evaluation}. Scoring is performed independently for each generated video. For each model and task level, every reported metric is the arithmetic mean of the corresponding per-video scores, rescaled to $[0,100]$. Thus, aggregate scores are computed from per-video quantities rather than reconstructed from marginal averages of the individual dimensions or scoring components.

\paragraph{Item Verdicts and Severity.}
Following VGI-Bench~\cite{he2026vgibenchprobingvisualintelligence}, we adopt an inverse-decay principle for violation-based evaluation. VGI-Bench applies an inverse-decay function, \(1/(1+x)\), to the number of repeated violations of an individual rubric item, reflecting the intuition that additional violations should have diminishing marginal impact once violations have already occurred. We adapt this principle to our task-conditioned evaluation setting, where checklist requirements differ in their procedural importance and consequences for successful task execution.

For each checklist item $i$ in a core evaluation dimension, we define:
\begin{equation}
y_i =
\begin{cases}
1, & \text{if item } i \text{ is judged \textsc{Yes}},\\
0, & \text{if item } i \text{ is judged \textsc{No} or \textsc{Unjudgeable}}.
\end{cases}
\end{equation}

Unlike uniform weighting across rubric items, laboratory procedures contain requirements with different consequences for successful execution. For example, omitting an essential operation or failing to realize a required final state is more consequential than violating a supplementary requirement. We therefore assign each item an expert-defined importance weight:
\begin{equation}
w_i =
\begin{cases}
1,    & \text{\emph{Critical}},\\
0.5,  & \text{\emph{Standard}},\\
0.25, & \text{\emph{Supplementary}}.
\end{cases}
\label{eq:weights}
\end{equation}

For each core dimension $d$, we define the accumulated weighted violation severity as
\begin{equation}
V_d = \sum_{i \in d} w_i (1-y_i),
\label{eq:severity}
\end{equation}
and map it to a normalized dimension score using inverse decay:
\begin{equation}
S_d = \frac{1}{1+V_d}.
\label{eq:dimension}
\end{equation}

This formulation preserves the inverse-decay treatment of accumulated violations while extending it to importance-aware, dimension-level evaluation. The weighting scheme provides a simple and interpretable severity hierarchy: one \emph{Critical} violation contributes the same severity as two \emph{Standard} violations or four \emph{Supplementary} violations. The inverse-decay mapping makes the first violations most consequential, while additional violations have progressively smaller marginal effects.

\paragraph{Core Visual Quality and Task Completion.}
The four core dimensions, namely \textit{Object Consistency}, \textit{Action Fidelity}, \textit{State Correctness}, and \textit{Physical Plausibility}, are combined through a geometric mean to obtain a balanced core score that penalizes any weak dimension:
\begin{equation}
S_{\mathrm{core}}
=
\left(
\prod_{d=1}^{K} S_d
\right)^{1/K},
\qquad K=4.
\label{eq:core}
\end{equation}
The geometric mean limits compensation across dimensions: a severe failure in one aspect of the generated procedure cannot be fully offset by strong performance elsewhere.

We separately measure whether the essential steps of the procedure are completed. Let $n_c$ denote the number of \emph{Critical} items in the four core dimensions that are not judged \textsc{Yes}. Scene-consistency and safety items are excluded from $n_c$, since they are handled by the validity constraints defined below. The completion factor is
\begin{equation}
C(n_c) =
\begin{cases}
1,   & n_c = 0,\\
0.5, & n_c = 1,\\
0,   & n_c \ge 2.
\end{cases}
\label{eq:completion}
\end{equation}
Hence, a single unmet critical requirement is treated as partial completion, whereas two or more indicate failure to complete the core procedure as specified.

\paragraph{Validity Constraints.}
Scene consistency and visual safety are treated as binary validity constraints because both are expected to hold under our TI2V setting. The input image largely fixes the scene configuration, while our prompts exclude unsafe actions. Violations therefore indicate exceptional model failures rather than ordinary quality variation, and are handled as veto conditions instead of graded scores. For $g\in\{\mathrm{scene},\mathrm{safety}\}$,
\begin{equation}
G_g \in \{0,1\},
\qquad
G_g = 0
\iff
\text{at least one item in dimension } g
\text{ is judged \textsc{No}}.
\label{eq:gates}
\end{equation}
All items within each validity dimension have equal veto power regardless of importance. An \textsc{Unjudgeable} item does not close the gate because missing evidence is not treated as a violation. Any explicit \textsc{No} in scene consistency or safety closes the corresponding gate. These gates can only penalize explicit validity violations and can never increase a video's score.

\paragraph{Overall Score.}
The final per-video score is the product of these four components:
\begin{equation}
S_{\mathrm{overall}} = S_{\mathrm{core}} \cdot C(n_c) \cdot G_{\mathrm{scene}} \cdot G_{\mathrm{safety}}.
\label{eq:overall1}
\end{equation}
This decomposition separates three distinct aspects of evaluation: $S_{\mathrm{core}}$ measures execution quality, $C(n_c)$ captures completion of essential procedural requirements, and $G_{\mathrm{scene}}$ and $G_{\mathrm{safety}}$ enforce validity of the resulting video. In particular, any failed validity constraint sets the overall score to zero regardless of the video's core visual quality or completion status.

\begin{table}[t]
    \centering
    \small
    \setlength{\tabcolsep}{8pt}
    \caption{Generators evaluated in the benchmark and their generation settings. $\dagger$ indicates commercial models.}
    \label{tab:evaluated_models}
    \resizebox{\linewidth}{!}{%
    \begin{tabular}{lcccccccc}
        \toprule
        \textbf{Organization} & \textbf{Model} & \textbf{Release} & \textbf{Scale} & \textbf{FPS} & \textbf{Resolution} & \textbf{Level 1/2 Frames} & \textbf{CFG} & \textbf{Steps} \\
        \midrule
        Alibaba & Wan 2.2 I2V A14B & 2025-07 & 14B & 16 & 480p & 65 / 193 & 3.5 & 40 \\
        Lightricks & LTX 2.3 & 2026-03 & 22B & 24 & 480p & 97 / 289 & 3.0 & 30 \\
        NVIDIA & Cosmos3 Nano & 2026-06 & 16B & 24 & 480p & 96 / 288 & 6.0 & 35 \\
        NVIDIA & Cosmos3 Super & 2026-06 & 64B & 24 & 480p & 96 / 288 & 6.0 & 35 \\
        ByteDance & LingBot Video & 2026-07 & 30B & 24 & 480p & 97 / 289 & 3.0 & 40 \\
        MiniMax & MiniMax~H3 & 2026-08 & 33B & 24 & 480p & 124 / 294 & -- & 30 \\
        ByteDance & $\text{Seedance 2.0}^\dagger$ & 2026-04 & -- & 24 & 480p & 96 / 288 & -- & -- \\
        Alibaba & $\text{Wan 3.0}^\dagger$ & 2026-08 & -- & 30 & 480p & 120 / 360 & -- & -- \\
        \bottomrule
    \end{tabular}}
\end{table}

\begin{table}[t]
    \centering
    \footnotesize
    \setlength{\tabcolsep}{16pt}
    \renewcommand{\arraystretch}{1}
    \caption{Benchmark scores by Gemini~3.7 Flash and GPT-5.6 Sol.}
    \label{tab:cross_evaluator_scores}
    \resizebox{\textwidth}{!}{
    \begin{tabular}{lcccccc}
        \toprule
        \multirow{2}{*}{\textbf{Model}}
        & \multicolumn{3}{c}{\textbf{Gemini 3.7 Flash}}
        & \multicolumn{3}{c}{\textbf{GPT-5.6 Sol}} \\
        \arrayrulecolor{sepgray}
        \cmidrule(lr){2-4}\cmidrule(lr){5-7}
        \arrayrulecolor{black}
        & \textbf{Level~1} & \textbf{Level~2} & \textbf{Overall}
        & \textbf{Level~1} & \textbf{Level~2} & \textbf{Overall} \\
        \midrule
        Wan 2.2
        & 30.1 & 11.3 & \hlcell{thirdfill}{20.6}
        & 38.5 & 13.6 & \hlcell{thirdfill}{25.9} \\
        LTX 2.3
        & 30.7 & 29.1 & \hlcell{thirdfill}{29.9}
        & 22.1 & 22.1 & \hlcell{thirdfill}{22.1} \\
        Cosmos3 Nano
        & 22.6 & 12.4 & \hlcell{thirdfill}{17.5}
        & 23.7 & 15.2 & \hlcell{thirdfill}{19.4} \\
        Cosmos3 Super
        & 28.7 & 20.5 & \hlcell{thirdfill}{24.6}
        & 30.0 & 15.7 & \hlcell{thirdfill}{22.8} \\
        LingBot Video
        & 30.7 & 21.6 & \hlcell{thirdfill}{26.1}
        & 20.0 & 19.1 & \hlcell{thirdfill}{19.5} \\
        MiniMax~H3
        & 48.7 & 53.8 & \hlcell{secondfill}{51.3}
        & \textbf{53.6} & 39.2 & \hlcell{bestfill}{\textbf{46.3}} \\
        $\text{Seedance 2.0}^\dagger$
        & 38.8 & 53.1 & \hlcell{thirdfill}{46.0}
        & 41.9 & \textbf{48.2} & \hlcell{secondfill}{45.1} \\
        $\text{Wan 3.0}^\dagger$
        & \textbf{60.2} & \textbf{60.0}
        & \hlcell{bestfill}{\textbf{60.1}}
        & 46.2 & 42.7 & \hlcell{thirdfill}{44.4} \\
        \bottomrule
    \end{tabular}}
\end{table}

\paragraph{Aggregation and Reporting.}
All quantities above are first computed independently at the video level. For each model--level pair, we then aggregate these per-video quantities by taking their arithmetic mean and multiply the resulting value by $100$ for reporting. Accordingly, the reported \textbf{Core} score is the mean of the per-video $S_{\mathrm{core}}$ values, while the \textbf{Completion} score is the mean of the per-video $C(n_c)$ values. Since \textbf{Scene} and \textbf{Safety} are binary constraints rather than scored dimensions, they are reported as gate validity rates, defined as the fractions of videos for which $G_{\mathrm{scene}}=1$ and $G_{\mathrm{safety}}=1$, respectively, under the corresponding evaluator.

The reported \textbf{Overall} score is the mean of the per-video $S_{\mathrm{overall}}$ values. Thus, each task contributes one video-level score to the final average, rather than first aggregating individual scoring components across tasks. When Overall is reported over the full benchmark rather than separately for each level, we refer to it as \textbf{Pooled Overall}, or simply \textbf{Overall} when the context is unambiguous. Pooled Overall is computed by directly pooling the per-video $S_{\mathrm{overall}}$ values from all \numtasks{} tasks and averaging them together. It is therefore not, in general, equal to the arithmetic mean of the Level~1 and Level~2 Overall scores, since the two levels may contain different numbers of tasks.

\begin{table}[t]
    \centering
    \setlength{\tabcolsep}{12pt}
    \caption{Model-level consistency between Gemini 3.7 Flash and GPT-5.6 Sol. Pearson and Spearman correlations are computed across the aggregate scores of the \nummodels{} evaluated generators.}
    \label{tab:vlm_model_consistency}
    \begin{tabular}{lcc}
        \toprule
        \textbf{Score}
        & \textbf{Pearson $r$}
        & \textbf{Spearman $\rho$} \\
        \midrule
        Level~1 & 0.783 & 0.611 \\
        Level~2 & 0.968 & 0.929 \\
        Overall & 0.923 & 0.714 \\
        \bottomrule
    \end{tabular}
\end{table}

\begin{table}[t]
    \centering
    \setlength{\tabcolsep}{17pt}
    \caption{Task-wise ranking agreement between Gemini 3.7 Flash and GPT-5.6 Sol, measured by Spearman correlation, Kendall correlation, and pairwise ordering agreement.}
    \label{tab:vlm_taskwise_consistency}
    \resizebox{\textwidth}{!}{
    \begin{tabular}{lccccc}
        \toprule
        \textbf{Split}
        & \textbf{\# Tasks}
        & \textbf{Mean $\rho$}
        & \textbf{Median $\rho$}
        & \textbf{Mean $\tau_b$}
        & \textbf{Pairwise Agree.} \\
        \midrule
        Level~1 & 86 & 0.279 & 0.280 & 0.269 & 67.0\% \\
        Level~2 & 81 & 0.415 & 0.539 & 0.388 & 74.7\% \\
        Overall & 167 & 0.345 & 0.426 & 0.327 & 70.8\% \\
        \bottomrule
    \end{tabular}}
\end{table}

\section{Experimental Details}

\subsection{Evaluated Models and Generation Settings}
\label{sec:model_details}

Table~\ref{tab:evaluated_models} summarizes the evaluated model releases and their native generation configurations. For each model, we select the supported frame count that most closely matches the nominal duration of approximately 4~s for Level~1 tasks and 12~s for Level~2 tasks. All videos are evaluated at their native temporal length, without frame interpolation, frame duplication, or temporal extension. Due to its native generation constraints, MiniMax~H3 produces slightly longer outputs for Level~1 tasks. To ensure a consistent setting, each video is generated at the aspect ratio of its task's initial frame, with the resolution set to the value closest to 480p that the corresponding model supports for that ratio; no model-specific prompt keywords or prompt-tuning strategies are used. Evaluation is restricted to the generated visual track, consistent with the benchmark's visually verifiable criteria.

\begin{figure}[t]
    \centering
    \includegraphics[width=1.0\linewidth]{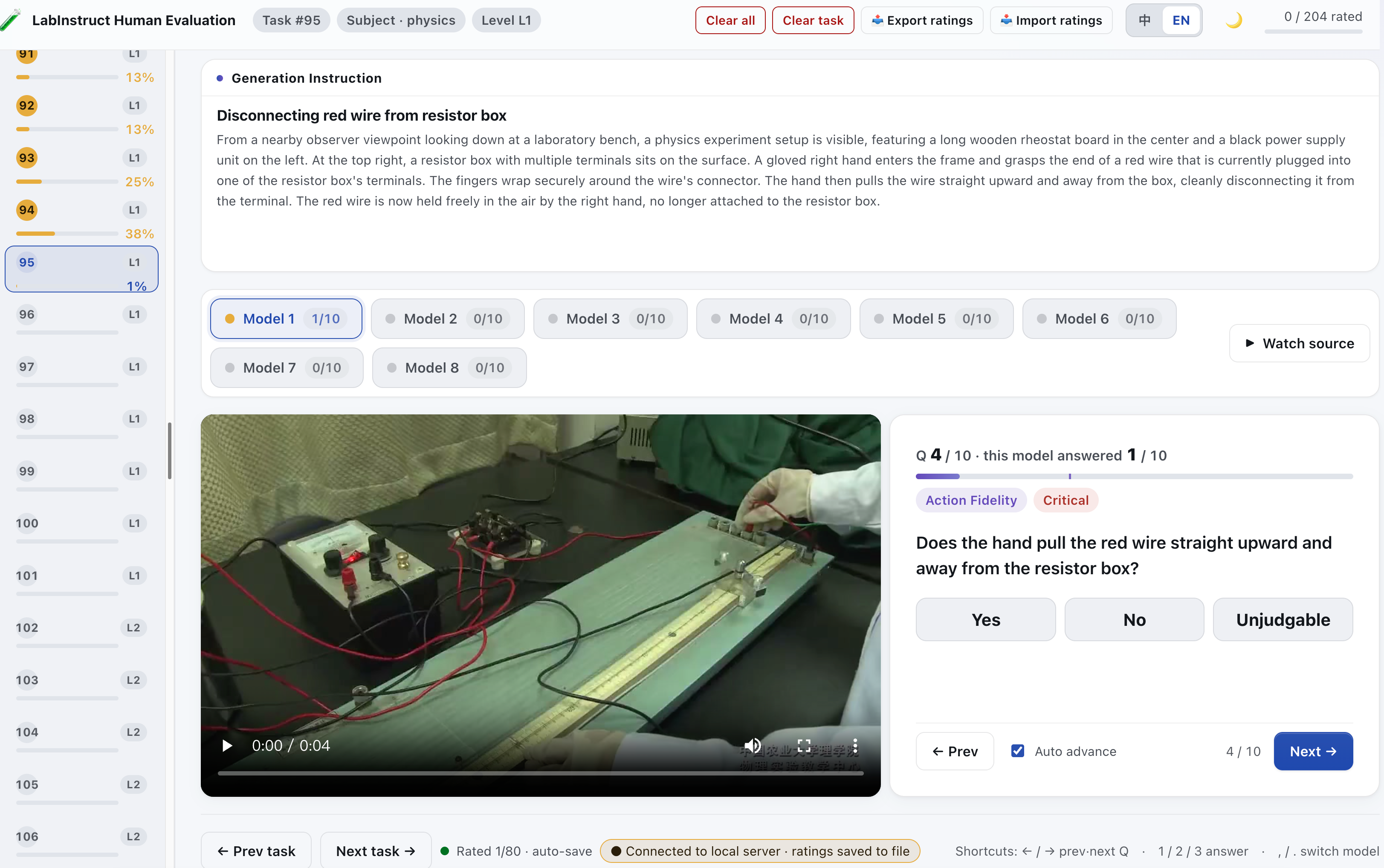}
    \caption{Human evaluation interface for task-specific assessment of generated videos.}
    \label{fig:human_eval}
\end{figure}

\subsection{Cross-Evaluator Robustness}
\label{sec:cross_evaluator_robustness}

\paragraph{Gemini~3.7 Flash Results.}
To examine the sensitivity of our benchmark results to the choice of VLM evaluator, we re-evaluate the full benchmark using Gemini~3.7 Flash~\cite{google2026gemini37flash} as an alternative judge, while keeping all other components of the evaluation protocol unchanged. Table~\ref{tab:cross_evaluator_scores} reports the results under both Gemini~3.7 Flash and GPT-5.6 Sol. Across the \nummodels{} generators, the Overall score averages 34.5 with Gemini~3.7 Flash and 30.7 with GPT-5.6 Sol. The highest model score reaches 60.1 under Gemini~3.7 Flash and 46.3 under GPT-5.6 Sol. Moreover, the same five generators remain below an Overall score of 30 under both evaluators. Although the absolute scores differ between the two judges, both evaluations place current generator performance well below benchmark saturation.
The relative performance of the strongest generators is also largely preserved across the two evaluators. MiniMax~H3, Seedance~2.0, and Wan~3.0 form the top-performing group under both Gemini~3.7 Flash and GPT-5.6 Sol, although their internal ordering differs. Beyond this top group, the two evaluators also produce broadly similar performance patterns across the remaining generators. Overall, changing the VLM evaluator affects score calibration more than the main comparative conclusions of the benchmark.

\paragraph{Model-Level Consistency.}
We first examine whether the two VLM evaluators produce consistent model-level results. For Level~1, Level~2, and Overall, we construct two vectors of aggregate scores across the \nummodels{} evaluated generators, one obtained with Gemini~3.7 Flash and the other with GPT-5.6 Sol. We compute Pearson correlation to quantify the linear association between the scores assigned by the two evaluators, and Spearman correlation to assess the consistency of the resulting generator rankings. As shown in Table~\ref{tab:vlm_model_consistency}, the Overall scores are strongly correlated across evaluators, with a Pearson correlation of $r=0.923$ and a Spearman correlation of $\rho=0.714$. The highest consistency is observed at Level~2 ($r=0.968$, $\rho=0.929$), where both the aggregate scores and generator rankings closely align across the two evaluators. Level~1 exhibits lower, though still positive, cross-evaluator correlations ($r=0.783$, $\rho=0.611$). Overall, the two evaluators produce similar model-level performance patterns, with particularly strong consistency on Level~2 and some differences in the finer-grained ordering of generators on Level~1 and in the overall results.

\paragraph{Task-Wise Ranking Agreement.}
We further evaluate the consistency of relative generator performance at the individual-task level. For each task, we consider generators with valid scores from both evaluators and compute Spearman's $\rho$ and Kendall's $\tau_b$ over their respective rankings. Degenerate rankings are excluded from the corresponding correlation statistic. We report the macro-average Spearman and Kendall correlations across tasks, together with the median task-wise Spearman correlation. We additionally measure pairwise agreement by considering generator pairs for which both evaluators produce a non-tied ordering and computing the fraction for which the two evaluators identify the same generator as higher scoring. The resulting pairwise agreement rates are likewise macro-averaged across the retained non-degenerate tasks.

Table~\ref{tab:vlm_taskwise_consistency} reports the task-wise ranking agreement between the two evaluators. Over the 167 tasks, the mean Spearman correlation is 0.345, the median Spearman correlation is 0.426, and the mean Kendall correlation is 0.327. The corresponding pairwise agreement rate is 70.8\%. At Level~2, the two evaluators achieve a mean Spearman correlation of 0.415, a median Spearman correlation of 0.539, a mean Kendall correlation of 0.388, and a pairwise agreement rate of 74.7\% across 81 tasks. For Level~1, the corresponding values are 0.279, 0.280, 0.269, and 67.0\% across 86 tasks. The higher correlation and pairwise-agreement values at Level~2 show that generator rankings are more consistent between the two evaluators on Level~2 tasks, whereas Level~1 exhibits greater differences in the task-wise relative ordering of generators.

\begin{table}[t]
\centering
\small
\setlength{\tabcolsep}{15pt}
\caption{Human--human agreement on pre-adjudication annotations. Fleiss' $\kappa$ measures chance-corrected agreement among all raters. Pairwise Agree. is averaged over all rater pairs.}
\begin{tabular}{lcc}
\toprule
\textbf{Comparison} & \textbf{Fleiss' $\kappa$} & \textbf{Pairwise Agree. (\%)} \\
\midrule
Human--Human & 0.57 & 80.7 \\
\bottomrule
\end{tabular}
\label{tab:human_human_agreement}
\end{table}

\begin{table}[t]
    \centering
    \small
    \setlength{\tabcolsep}{17pt}
    \caption{Task-wise ranking agreement between Human and VLM evaluation. \# Tasks counts the tasks whose generator ranking is non-degenerate under both evaluators. Fully tied rankings are excluded from Spearman $\rho$, Kendall's $\tau_b$, and the pairwise agreement rate alike.}
    \label{tab:human_vlm_taskwise_consistency}
    \resizebox{\textwidth}{!}{%
    \begin{tabular}{lccccc}
        \toprule
        \textbf{Split} & \textbf{\# Tasks} & \textbf{Mean $\rho$} & \textbf{Median $\rho$} & \textbf{Mean $\tau_b$} & \textbf{Pairwise Agree.} \\
        \midrule
        Level~1 & 79 & 0.351 & 0.378 & 0.331 & 72.9\% \\
        Level~2 & 60 & 0.380 & 0.432 & 0.364 & 73.0\% \\
        Overall & 139 & 0.364 & 0.378 & 0.345 & 72.9\% \\
        \bottomrule
    \end{tabular}}
\end{table}

\begin{table}[t]
    \centering
    \small
    \setlength{\tabcolsep}{5pt}
    \renewcommand{\arraystretch}{1}
    \caption{Sensitivity of pooled model scores under alternative core-score normalization, completion-threshold formulations, and strict completion criteria. Higher is better. Best values are bolded, subscripts indicate within-row rank, and $\dagger$ denotes commercial models.}
    \label{tab:scoring_sensitivity}
    \resizebox{\linewidth}{!}{%
    \begin{tabular}{llcccccccc}
        \toprule
        \textbf{Criterion} & \textbf{Variant} & \hlhead{\colhead{\textbf{Wan}\\ \textbf{2.2}}} & \hlhead{\colhead{\textbf{LTX}\\ \textbf{2.3}}} & \hlhead{\colhead{\textbf{Cosmos3}\\ \textbf{Nano}}} & \hlhead{\colhead{\textbf{Cosmos3}\\ \textbf{Super}}} & \hlhead{\colhead{\textbf{LingBot}\\ \textbf{Video}}} & \hlhead{\colhead{\textbf{MiniMax}\\ \textbf{H3}}} & \hlhead{\colhead{\textbf{Seedance}\\ \textbf{2.0}$^\dagger$}} & \hlhead{\colhead{\textbf{Wan}\\ \textbf{3.0}$^\dagger$}} \\
        \midrule
        \multirow{2}{*}{$S_{\mathrm{core}}$} & Original & 68.0$_4$ & 66.0$_5$ & 63.0$_8$ & 65.2$_6$ & 64.9$_7$ & 79.8$_2$ & 78.7$_3$ & \textbf{81.0$_1$} \\
        & \textit{Norm.} & 82.8$_4$ & 80.9$_5$ & 79.4$_8$ & 80.8$_6$ & 80.4$_7$ & 89.7$_2$ & 89.1$_3$ & \textbf{90.4$_1$} \\
        \midrule
        \multirow{2}{*}{Comp.} & Original & 30.6$_3$ & 26.2$_4$ & 21.5$_7$ & 24.7$_5$ & 22.7$_6$ & \textbf{51.9$_1$} & 49.5$_2$ & \textbf{51.9$_1$} \\
        & \textit{Relative} & 32.8$_4$ & 28.1$_5$ & 22.5$_8$ & 26.4$_6$ & 24.2$_7$ & \textbf{54.6$_1$} & 51.4$_3$ & 54.4$_2$ \\
        \midrule
        \multirow{5}{*}{Overall} & Original & 25.9$_4$ & 22.1$_6$ & 19.4$_8$ & 22.8$_5$ & 19.5$_7$ & \textbf{46.3$_1$} & 45.1$_2$ & 44.4$_3$ \\
        & + \textit{Norm.} $S_{\mathrm{core}}$ & 26.9$_4$ & 22.8$_6$ & 19.9$_8$ & 23.5$_5$ & 20.5$_7$ & \textbf{47.2$_1$} & 46.2$_2$ & 45.4$_3$ \\
        & + \textit{Relative} Comp. & 27.5$_4$ & 23.1$_6$ & 20.0$_8$ & 23.7$_5$ & 20.5$_7$ & \textbf{47.9$_1$} & 46.4$_2$ & 45.8$_3$ \\
        & + \textit{Both} & 29.0$_4$ & 24.1$_6$ & 20.8$_8$ & 24.6$_5$ & 21.8$_7$ & \textbf{49.3$_1$} & 47.9$_2$ & 47.2$_3$ \\
        & \textit{Strict} Overall & 20.4$_4$ & 17.4$_6$ & 15.9$_7$ & 18.8$_5$ & 13.3$_8$ & \textbf{40.3$_1$} & 39.1$_2$ & 38.3$_3$ \\
        \bottomrule
    \end{tabular}}
\end{table}

\begin{table}[t]
    \centering
    \small
    \setlength{\tabcolsep}{12pt}
    \renewcommand{\arraystretch}{1}
    \caption{Sensitivity of Overall scores to evaluator sampling rate. Bold indicates the highest score across 2, 4, and 8 FPS for each model under each evaluator. $\dagger$ indicates commercial models.}
    \label{tab:fps}
    \begin{tabular*}{\textwidth}{@{\extracolsep{\fill}}lcccccc@{\extracolsep{0pt}}}
        \toprule
        \multirow{2}{*}{\textbf{Model}}
        & \multicolumn{3}{c}{\textbf{GPT-5.6 Sol}}
        & \multicolumn{3}{c}{\textbf{Gemini 3.7 Flash}} \\
        \arrayrulecolor{sepgray}\cmidrule(lr){2-4}\cmidrule(lr){5-7}\arrayrulecolor{black}
        & \textbf{FPS 2} & \textbf{FPS 4} & \textbf{FPS 8}
        & \textbf{FPS 2} & \textbf{FPS 4} & \textbf{FPS 8} \\
        \midrule
        Wan 2.2 & 13.3 & 25.9 & \textbf{31.3} & \textbf{24.8} & 20.6 & 24.4 \\
        LTX 2.3 & 21.1 & 22.1 & \textbf{23.5} & 31.5 & 29.9 & \textbf{35.3} \\
        Cosmos3 Nano & 13.9 & 19.4 & \textbf{25.2} & 17.5 & 17.5 & \textbf{18.8} \\
        Cosmos3 Super & 18.8 & 22.8 & \textbf{28.8} & 21.9 & 24.6 & \textbf{26.1} \\
        LingBot Video & 17.2 & 19.5 & \textbf{29.1} & 25.9 & 26.1 & \textbf{26.9} \\
        MiniMax~H3 & 41.7 & 46.3 & \textbf{48.6} & 50.4 & 51.3 & \textbf{51.7} \\
        $\text{Seedance 2.0}^\dagger$ & 30.2 & 45.1 & \textbf{46.6} & \textbf{47.5} & 46.0 & 46.0 \\
        $\text{Wan 3.0}^\dagger$ & 24.9 & 44.4 & \textbf{48.9} & 60.5 & 60.1 & \textbf{61.1} \\
        \bottomrule
    \end{tabular*}
\end{table}

\subsection{Human Evaluation Details}
\label{sec:human_evaluation_details}

\paragraph{Evaluation Protocol.}
Human judgments serve as the reference for assessing the reliability of the automatic evaluator in Sec.~\ref{sec:main_results}. We conduct human evaluation over the full evaluation set, covering every generated video and all applicable task-specific checklist items. Each video is independently evaluated by four qualified raters using the same checklist questions and evaluation criteria as the VLM evaluator. The four raters include two master's-level researchers and two doctoral-level researchers. We develop a dedicated annotation interface (Figure~\ref{fig:human_eval}) that presents the task instruction, task metadata, generated video, and one checklist question at a time. Raters may consult the source video only to clarify task context or object identity; whether a checklist item is satisfied must be decided solely from evidence in the generated video. To reduce model-specific bias, generator identities are anonymized throughout annotation. The interface also displays each item's evaluation dimension and highlights critical items when applicable.

Before formal annotation, all raters complete a calibration session using representative examples. For each checklist item, raters assign one of three labels: \textsc{Yes}, \textsc{No}, or \textsc{Unjudgeable}. Unanimous judgments are accepted directly, while disagreements are resolved through expert adjudication. Cases involving subtle state transitions, liquid handling, transparent objects, brief contacts, or safety-critical actions receive additional review when needed.

Human evaluation follows the same core judgment principles as the VLM evaluator:
\begin{itemize}

\item \textbf{Evidence-based Judgment.}
Raters base each verdict only on evidence visible in the generated video and do not infer correctness from the task description, expected procedure, final-outcome plausibility, or source-video execution.

\item \textbf{Yes/No Criteria.}
A checklist item is marked \textsc{Yes} only when the generated video clearly demonstrates that the required action, state, or condition is satisfied, and \textsc{No} when it clearly shows that the requirement is absent, violated, or incorrectly performed.

\item \textbf{Unjudgeable Under Limited Visibility.}
An item is marked \textsc{Unjudgeable} when the available visual evidence is insufficient to determine whether the requirement is satisfied, such as under occlusion, severe blur, objects leaving the frame, or visual artifacts. Failure to observe an event is not, by itself, treated as evidence that the event did not occur or as grounds for a \textsc{No} verdict under this evidence-only policy.

\item \textbf{Temporal and Conditional Relations.}
For checklist items involving temporal order or conditional dependencies, raters assess the required relation when all relevant events are reliably observable. A clearly missing event or an observed ordering that violates the checklist is marked \textsc{No}. If event occurrence or ordering cannot be determined reliably, the item is marked \textsc{Unjudgeable} to preserve the distinction between observed failure and missing evidence in the final annotations.

\end{itemize}

\paragraph{Human Inter-Rater Reliability.}
We assess the reliability of the human judgments using the independent annotations collected before expert adjudication. Each checklist item is rated by all four human evaluators using the same three categorical labels, \textsc{Yes}, \textsc{No}, and \textsc{Unjudgeable}. As shown in Table~\ref{tab:human_human_agreement}, the four raters achieve a Fleiss' $\kappa$ of 0.57 and a mean pairwise agreement of 80.7\% over all six rater pairs. These results indicate moderate chance-corrected inter-rater reliability together with high raw agreement prior to adjudication.

\paragraph{Task-Wise Ranking Agreement.}
We further examine whether human and VLM evaluations induce consistent generator rankings at the individual task level. For each task, we independently rank the \nummodels{} generators using their human-based and VLM-based overall scores, then compare the two rankings using Spearman $\rho$ and Kendall's $\tau_b$. We also compute pairwise agreement as the fraction of non-tied generator pairs for which the two evaluators identify the same generator as better, and macro-average this quantity across tasks. Tasks with fully tied rankings under either evaluator are excluded from the corresponding ranking statistics, leaving 79 Level~1 tasks and 60 Level~2 tasks for comparison. As shown in Table~\ref{tab:human_vlm_taskwise_consistency}, the two evaluators exhibit moderate but consistent agreement within individual tasks. Across all 139 valid tasks, the mean Spearman correlation is $\rho=0.364$, the median Spearman correlation is 0.378, and the mean Kendall correlation is $\tau_b=0.345$. At the pairwise level, human and VLM evaluations agree on 72.9\% of non-tied generator comparisons, indicating that they identify the stronger generator in nearly three quarters of pairwise decisions. Agreement is also stable across the two strata, with pairwise agreement rates of 72.9\% for Level~1 and 73.0\% for Level~2, while Level~2 shows slightly higher mean Spearman and Kendall correlations. These results suggest that the VLM evaluator preserves a meaningful portion of the task-specific relative performance signal, although exact rankings remain less stable at the individual task level than after aggregation across the full benchmark.

\begin{table}[t]
    \centering
    \setlength{\tabcolsep}{6pt}
    \renewcommand{\arraystretch}{1}
    \caption{Overall score by operation domain and task level. $\dagger$ denotes commercial models.}
    \label{tab:domain_breakdown}
    \resizebox{\linewidth}{!}{%
    \begin{tabular}{llcccccccc}
        \toprule
        \textbf{Domain} & \textbf{Level} & \hlhead{\colhead{\textbf{Wan}\\ \textbf{2.2}}} & \hlhead{\colhead{\textbf{LTX}\\ \textbf{2.3}}} & \hlhead{\colhead{\textbf{Cosmos3}\\ \textbf{Nano}}} & \hlhead{\colhead{\textbf{Cosmos3}\\ \textbf{Super}}} & \hlhead{\colhead{\textbf{LingBot}\\ \textbf{Video}}} & \hlhead{\colhead{\textbf{MiniMax}\\ \textbf{H3}}} & \hlhead{\colhead{\textbf{Seedance}\\ \textbf{2.0}$^\dagger$}} & \hlhead{\colhead{\textbf{Wan}\\ \textbf{3.0}$^\dagger$}} \\
        \midrule
        \multirow{3}{*}{Container Operations}         & Level~1 & 76.0 & 48.8 & 71.4 & 34.2 & 24.0 & \textbf{84.3} & 70.0 & 52.9 \\
         & Level~2 & 16.7 & 20.0 & 9.3 & 15.9 & 20.3 & 29.3 & \textbf{51.7} & 36.2 \\
         & Overall & \hlcellwide{thirdfill}{26.6} & \hlcellwide{thirdfill}{24.8} & \hlcellwide{thirdfill}{19.7} & \hlcellwide{thirdfill}{18.9} & \hlcellwide{thirdfill}{20.9} & \hlcellwide{thirdfill}{38.4} & \hlcellwide{bestfill}{\textbf{54.7}} & \hlcellwide{secondfill}{39.0} \\
        \midrule
        \multirow{3}{*}{Filtration/Separation}         & Level~1 & \textbf{66.6} & 47.3 & 0.0 & 0.0 & 26.6 & 33.3 & 0.0 & 0.0 \\
         & Level~2 & 28.4 & 28.4 & 54.7 & 0.0 & 38.0 & 38.0 & \textbf{60.0} & \textbf{60.0} \\
         & Overall & \hlcellwide{bestfill}{\textbf{42.7}} & \hlcellwide{thirdfill}{35.5} & \hlcellwide{thirdfill}{34.2} & \hlcellwide{thirdfill}{0.0} & \hlcellwide{thirdfill}{33.8} & \hlcellwide{thirdfill}{36.3} & \hlcellwide{secondfill}{37.5} & \hlcellwide{secondfill}{37.5} \\
        \midrule
        \multirow{3}{*}{Heating/Cooling}         & Level~1 & 0.0 & 0.0 & 0.0 & 0.0 & 0.0 & 50.0 & 0.0 & \textbf{71.0} \\
         & Level~2 & 0.0 & 0.0 & 20.0 & 0.0 & 8.4 & 40.0 & \textbf{88.4} & 27.9 \\
         & Overall & \hlcellwide{thirdfill}{0.0} & \hlcellwide{thirdfill}{0.0} & \hlcellwide{thirdfill}{14.2} & \hlcellwide{thirdfill}{0.0} & \hlcellwide{thirdfill}{6.0} & \hlcellwide{secondfill}{42.8} & \hlcellwide{bestfill}{\textbf{63.1}} & \hlcellwide{thirdfill}{40.2} \\
        \midrule
        \multirow{3}{*}{Instrument Operations}         & Level~1 & 54.9 & 18.0 & 25.8 & 32.4 & 21.2 & \textbf{55.2} & 49.1 & 52.8 \\
         & Level~2 & 16.3 & 25.4 & 19.8 & 21.5 & 21.2 & 57.3 & \textbf{57.7} & 47.4 \\
         & Overall & \hlcellwide{thirdfill}{36.2} & \hlcellwide{thirdfill}{21.6} & \hlcellwide{thirdfill}{22.9} & \hlcellwide{thirdfill}{27.1} & \hlcellwide{thirdfill}{21.2} & \hlcellwide{bestfill}{\textbf{56.2}} & \hlcellwide{secondfill}{53.2} & \hlcellwide{thirdfill}{50.2} \\
        \midrule
        \multirow{3}{*}{Liquid Handling}         & Level~1 & 15.2 & 16.7 & 12.8 & 21.6 & 13.6 & 31.2 & 28.7 & \textbf{38.8} \\
         & Level~2 & 11.1 & 12.2 & 8.5 & 5.8 & 20.2 & 18.4 & 29.9 & \textbf{33.5} \\
         & Overall & \hlcellwide{thirdfill}{12.8} & \hlcellwide{thirdfill}{14.1} & \hlcellwide{thirdfill}{10.3} & \hlcellwide{thirdfill}{12.5} & \hlcellwide{thirdfill}{17.4} & \hlcellwide{thirdfill}{23.8} & \hlcellwide{secondfill}{29.4} & \hlcellwide{bestfill}{\textbf{35.7}} \\
        \midrule
        \multirow{3}{*}{Measurement}         & Level~1 & 33.9 & 21.3 & 7.8 & 37.0 & 28.6 & 57.6 & 41.9 & \textbf{64.5} \\
         & Level~2 & 12.7 & 29.6 & 5.8 & 11.7 & 0.0 & 43.0 & 46.9 & \textbf{56.7} \\
         & Overall & \hlcellwide{thirdfill}{23.0} & \hlcellwide{thirdfill}{25.6} & \hlcellwide{thirdfill}{6.8} & \hlcellwide{thirdfill}{24.0} & \hlcellwide{thirdfill}{13.8} & \hlcellwide{secondfill}{50.1} & \hlcellwide{thirdfill}{44.4} & \hlcellwide{bestfill}{\textbf{60.4}} \\
        \midrule
        \multirow{3}{*}{Solid Handling}         & Level~1 & 31.3 & 25.4 & 38.3 & 38.7 & 18.6 & \textbf{71.5} & 49.1 & 30.4 \\
         & Level~2 & 6.0 & 25.0 & 7.6 & 18.4 & 13.5 & 42.4 & \textbf{44.4} & 40.4 \\
         & Overall & \hlcellwide{thirdfill}{16.2} & \hlcellwide{thirdfill}{25.2} & \hlcellwide{thirdfill}{20.0} & \hlcellwide{thirdfill}{26.6} & \hlcellwide{thirdfill}{15.5} & \hlcellwide{bestfill}{\textbf{54.2}} & \hlcellwide{secondfill}{46.3} & \hlcellwide{thirdfill}{36.3} \\
        \bottomrule
    \end{tabular}}
\end{table}

\begin{table}[t]
    \centering
    \setlength{\tabcolsep}{6pt}
    \renewcommand{\arraystretch}{1}
    \caption{Overall score by scientific discipline and task level. $\dagger$ denotes commercial models.}
    \label{tab:discipline_breakdown}
    \resizebox{\linewidth}{!}{%
    \begin{tabular}{llcccccccc}
        \toprule
        \textbf{Discipline} & \textbf{Level} & \hlhead{\colhead{\textbf{Wan}\\ \textbf{2.2}}} & \hlhead{\colhead{\textbf{LTX}\\ \textbf{2.3}}} & \hlhead{\colhead{\textbf{Cosmos3}\\ \textbf{Nano}}} & \hlhead{\colhead{\textbf{Cosmos3}\\ \textbf{Super}}} & \hlhead{\colhead{\textbf{LingBot}\\ \textbf{Video}}} & \hlhead{\colhead{\textbf{MiniMax}\\ \textbf{H3}}} & \hlhead{\colhead{\textbf{Seedance}\\ \textbf{2.0}$^\dagger$}} & \hlhead{\colhead{\textbf{Wan}\\ \textbf{3.0}$^\dagger$}} \\
        \midrule
        \multirow{3}{*}{Agronomy}         & Level~1 & 27.0 & 53.2 & 31.5 & 34.2 & 18.4 & \textbf{57.0} & 36.4 & 34.2 \\
         & Level~2 & 5.9 & 13.4 & 17.5 & 18.4 & 3.2 & \textbf{49.3} & 38.3 & 35.3 \\
         & Overall & \hlcellwide{thirdfill}{15.1} & \hlcellwide{thirdfill}{30.7} & \hlcellwide{thirdfill}{23.6} & \hlcellwide{thirdfill}{25.2} & \hlcellwide{thirdfill}{9.8} & \hlcellwide{bestfill}{\textbf{52.7}} & \hlcellwide{secondfill}{37.5} & \hlcellwide{thirdfill}{34.8} \\
        \midrule
        \multirow{3}{*}{Biology}         & Level~1 & 21.6 & 16.9 & 21.6 & 19.5 & 13.7 & \textbf{28.7} & 25.8 & 22.8 \\
         & Level~2 & 11.5 & 14.2 & 7.9 & 7.6 & 20.1 & 9.0 & \textbf{36.2} & 26.0 \\
         & Overall & \hlcellwide{thirdfill}{16.1} & \hlcellwide{thirdfill}{15.4} & \hlcellwide{thirdfill}{14.1} & \hlcellwide{thirdfill}{13.0} & \hlcellwide{thirdfill}{17.2} & \hlcellwide{thirdfill}{18.0} & \hlcellwide{bestfill}{\textbf{31.5}} & \hlcellwide{secondfill}{24.5} \\
        \midrule
        \multirow{3}{*}{Chemistry}         & Level~1 & 60.5 & 17.7 & 39.4 & 28.7 & 28.7 & \textbf{69.7} & 45.4 & 53.9 \\
         & Level~2 & 18.2 & 24.4 & 18.9 & 13.5 & 24.9 & 41.0 & 48.2 & \textbf{49.7} \\
         & Overall & \hlcellwide{thirdfill}{39.8} & \hlcellwide{thirdfill}{21.0} & \hlcellwide{thirdfill}{29.4} & \hlcellwide{thirdfill}{21.2} & \hlcellwide{thirdfill}{26.8} & \hlcellwide{bestfill}{\textbf{55.6}} & \hlcellwide{thirdfill}{46.8} & \hlcellwide{secondfill}{51.8} \\
        \midrule
        \multirow{3}{*}{Materials Science}         & Level~1 & 28.4 & 48.4 & 40.0 & 60.0 & 20.0 & 60.0 & \textbf{80.0} & 28.4 \\
         & Level~2 & 5.2 & 25.0 & 12.5 & 4.7 & 21.3 & 34.3 & \textbf{37.5} & 30.2 \\
         & Overall & \hlcellwide{thirdfill}{14.1} & \hlcellwide{thirdfill}{34.0} & \hlcellwide{thirdfill}{23.0} & \hlcellwide{thirdfill}{26.0} & \hlcellwide{thirdfill}{20.8} & \hlcellwide{secondfill}{44.1} & \hlcellwide{bestfill}{\textbf{53.8}} & \hlcellwide{thirdfill}{29.5} \\
        \midrule
        \multirow{3}{*}{Physics}         & Level~1 & 36.8 & 16.6 & 10.6 & 31.4 & 17.9 & 54.1 & 44.3 & \textbf{58.2} \\
         & Level~2 & 16.8 & 29.0 & 17.4 & 25.0 & 19.8 & 56.9 & \textbf{63.3} & 55.5 \\
         & Overall & \hlcellwide{thirdfill}{27.7} & \hlcellwide{thirdfill}{22.2} & \hlcellwide{thirdfill}{13.7} & \hlcellwide{thirdfill}{28.5} & \hlcellwide{thirdfill}{18.7} & \hlcellwide{secondfill}{55.4} & \hlcellwide{thirdfill}{52.9} & \hlcellwide{bestfill}{\textbf{57.0}} \\
        \bottomrule
    \end{tabular}}
\end{table}

\begin{table}[t]
    \centering
    \setlength{\tabcolsep}{6pt}
    \renewcommand{\arraystretch}{1}
    \caption{Overall score by experiment viewpoint and task level. $\dagger$ denotes commercial models.}
    \label{tab:viewpoint_breakdown}
    \resizebox{\linewidth}{!}{%
    \begin{tabular}{llcccccccc}
        \toprule
        \textbf{Viewpoint} & \textbf{Level} & \hlhead{\colhead{\textbf{Wan}\\ \textbf{2.2}}} & \hlhead{\colhead{\textbf{LTX}\\ \textbf{2.3}}} & \hlhead{\colhead{\textbf{Cosmos3}\\ \textbf{Nano}}} & \hlhead{\colhead{\textbf{Cosmos3}\\ \textbf{Super}}} & \hlhead{\colhead{\textbf{LingBot}\\ \textbf{Video}}} & \hlhead{\colhead{\textbf{MiniMax}\\ \textbf{H3}}} & \hlhead{\colhead{\textbf{Seedance}\\ \textbf{2.0}$^\dagger$}} & \hlhead{\colhead{\textbf{Wan}\\ \textbf{3.0}$^\dagger$}} \\
        \midrule
        \multirow{3}{*}{First-Person}         & Level~1 & 31.9 & 15.2 & 11.3 & 20.8 & 15.3 & \textbf{44.2} & 34.7 & 41.3 \\
         & Level~2 & 11.6 & 24.7 & 13.2 & 12.5 & 21.2 & 26.2 & 46.0 & \textbf{46.6} \\
         & Overall & \hlcellwide{thirdfill}{20.7} & \hlcellwide{thirdfill}{20.5} & \hlcellwide{thirdfill}{12.4} & \hlcellwide{thirdfill}{16.2} & \hlcellwide{thirdfill}{18.6} & \hlcellwide{thirdfill}{34.3} & \hlcellwide{secondfill}{41.0} & \hlcellwide{bestfill}{\textbf{44.2}} \\
        \midrule
        \multirow{3}{*}{Nearby Observer}         & Level~1 & 41.3 & 25.0 & 28.9 & 33.9 & 21.9 & \textbf{57.6} & 45.0 & 48.3 \\
         & Level~2 & 14.8 & 20.7 & 16.3 & 17.6 & 18.0 & 46.4 & \textbf{49.4} & 40.5 \\
         & Overall & \hlcellwide{thirdfill}{28.5} & \hlcellwide{thirdfill}{22.9} & \hlcellwide{thirdfill}{22.8} & \hlcellwide{thirdfill}{26.0} & \hlcellwide{thirdfill}{20.0} & \hlcellwide{bestfill}{\textbf{52.2}} & \hlcellwide{secondfill}{47.1} & \hlcellwide{thirdfill}{44.5} \\
        \bottomrule
    \end{tabular}}
\end{table}

\begin{table}[ht]
\centering
\small
\setlength{\tabcolsep}{12pt}
\caption{Best-of-three sampling on the balanced 40-task subset.}
\label{tab:best_of_three}
\resizebox{0.8\linewidth}{!}{
\begin{tabular}{lcccc}
\toprule
\textbf{Model}  & \textbf{Original}  & \textbf{Best-of-3}  & \textbf{Gain}  & \textbf{Original Best (\%)} \\  \midrule
MiniMax~H3  & 55.3 & 57.8 & +2.5 & 95.0 \\
Seedance 2.0$^\dagger$  & 52.8 & 66.8 & +14.0 & 80.0 \\
Wan~3.0$^\dagger$  & 53.7 & 64.8 & +11.1 & 85.0 \\
\bottomrule
\end{tabular}
}
\end{table}

\subsection{Additional Analyses}
\label{sec:additional_analyses}

\subsubsection{Scoring Sensitivity and Strict Overall}
\label{sec:completion_sensitivity}
The benchmark score combines two main components: a core score, $S_{\mathrm{core}}$, that aggregates weighted violations across the core dimensions, and a completion component that accounts for whether Critical requirements are satisfied. We examine whether the resulting model comparison is sensitive to the specific formulation of either component. Table~\ref{tab:scoring_sensitivity} reports pooled scores over tasks from both levels under the original formulation and several alternative variants.

\paragraph{Scoring Variants.}
For $S_{\mathrm{core}}$, the \emph{Norm.} variant normalizes the accumulated violation severity in each dimension by the total available weight:
\begin{equation}
\bar V_d = \frac{\sum_{i \in d} w_i(1-y_i)} {\sum_{i \in d} w_i}.
\end{equation}
Thus, severity reflects the fraction of weighted requirements violated rather than the absolute amount of checklist weight. For the Comp. metric, the \emph{Relative} variant expresses the partial-credit threshold relative to the number of Critical requirements. Let $N_{\mathrm{crit}}$ denote the number of Critical items in the four core dimensions and $n_c$ the number of those items not judged \textsc{Yes}. We define
\begin{equation}
\tilde{C}(n_c) =
\begin{cases}
1, & n_c = 0,\\
0.5, & 1 \le n_c \le 0.2\,N_{\mathrm{crit}},\\
0, & n_c > 0.2\,N_{\mathrm{crit}},
\end{cases}
\end{equation}
so that partial credit depends on the fraction of unmet Critical requirements rather than on an absolute count. In the Overall block, \emph{+ Norm.\ $S_{\mathrm{core}}$} and \emph{+ Relative Comp.} apply these two changes separately, while \emph{+ Both} applies them jointly. The \emph{Strict} variant instead removes partial completion credit altogether, assigning zero completion whenever any Critical requirement is unmet:
\begin{equation}
\hat{C}(n_c) =
\begin{cases}
1,   & n_c = 0,\\
0,   & n_c \ge 1.
\end{cases}
\label{eq:strict_completion}
\end{equation}
Consequently, a failure in any single Critical item is equivalent to failure of the entire task, which demands that the model be capable of end-to-end, error-free execution.

\paragraph{Ranking Robustness Across Scoring Variants.}
Alternative formulations affect score magnitudes more than model ordering. Normalizing $S_{\mathrm{core}}$ increases its value by 9.4--16.4 points while leaving the ranking unchanged, and changes Overall by only 0.5--1.1 points. Using the relative completion rule $\tilde{C}$ increases Overall by 0.6--1.6 points, while applying both modifications increases it by 1.4--3.1 points. Nevertheless, the Overall ranking is identical under \emph{Original}, \emph{+ Norm.\ $S_{\mathrm{core}}$}, \emph{+ Relative Comp.}, and \emph{+ Both}. Under the strict rule $\hat{C}$, the first six positions remain unchanged, with only LingBot~Video and Cosmos3~Nano exchanging the final two positions. These results show that the model ordering is robust to the formulation of both $S_{\mathrm{core}}$ and completion.

\paragraph{Completion Affects the Ordering of Leading Models.}
The component scores nevertheless emphasize different aspects of performance. Under the original $S_{\mathrm{core}}$, Wan~3.0 ranks first, followed by MiniMax~H3 and Seedance~2.0, whereas the Overall score orders them as MiniMax~H3, Seedance~2.0, and Wan~3.0. Thus, incorporating completion changes the relative ordering of the leading models. At the same time, this ordering remains unchanged across $\tilde{C}$, normalized $S_{\mathrm{core}}$, and their joint variant, indicating that the result is not driven by a particular scoring formulation.

\subsubsection{Sensitivity to Evaluator Sampling Rate}

Table~\ref{tab:fps} evaluates the effect of frame sampling rate by re-scoring the full benchmark at 2, 4, and 8 FPS under both GPT-5.6 Sol and Gemini~3.7 Flash, with all other evaluation settings unchanged. All reported scores pool tasks from both levels. Gemini~3.7 Flash is relatively stable across sampling rates: Wan~3.0, MiniMax~H3, and Seedance~2.0 remain the top three models at all three rates, and the mean score across models varies only from 34.5 to 36.3. GPT-5.6 Sol is more sensitive to sampling density, with the mean score increasing from 22.6 at 2~FPS to 30.7 at 4~FPS and 35.2 at 8~FPS. The increase is particularly pronounced for Wan~3.0 and Seedance~2.0, whereas MiniMax~H3 changes more modestly. Despite these score shifts, the same three models form the leading group across all sampling rates. We use 4~FPS as the default setting in the main experiments to balance evaluation fidelity and computational cost. Compared with 2~FPS, 4~FPS substantially improves temporal coverage, particularly for GPT-5.6 Sol, while avoiding the additional inference cost and latency of processing twice as many sampled frames at 8~FPS. Since increasing the rate to 8~FPS mainly changes absolute scores while leaving the leading group largely unchanged, 4~FPS provides a practical operating point for full-benchmark evaluation.

\subsubsection{Results by Operation Domain, Discipline, and Viewpoint}

\paragraph{Performance by Operation Domain.}
Table~\ref{tab:domain_breakdown} reveals clear operation-domain specialization among the strongest generators. MiniMax~H3 achieves the highest average score in instrument operations and solid handling, Seedance~2.0 leads container operations and heating/cooling, and Wan~3.0 performs best in liquid handling and measurement. Filtration/separation is the only domain led by a model outside this overall top group, with Wan~2.2 achieving the highest score of 42.7. Overall, the leading models exhibit complementary strengths across operation domains rather than a uniformly dominant capability profile, and no single generator performs best across the full range of laboratory operations. Cross-level differences are interpreted descriptively, since Level~1 and Level~2 contain independently curated task sets rather than controlled variants of the same tasks.

\paragraph{Performance by Scientific Discipline.}
Table~\ref{tab:discipline_breakdown} presents a similar pattern of complementary strengths across scientific disciplines. MiniMax~H3 achieves the highest average score in agronomy and chemistry, Seedance~2.0 leads biology and materials science, and Wan~3.0 performs best in physics. The performance gap between the leading models varies substantially by discipline: MiniMax~H3 leads agronomy by 15.2 points, whereas Wan~3.0 exceeds MiniMax~H3 by only 1.6 points in physics. Thus, model differentiation is much stronger in some disciplines than in others. Across all 5 disciplines, the highest scores are achieved by MiniMax~H3, Seedance~2.0, or Wan~3.0, but no single model leads every discipline. The discipline-level results therefore reveal distinct capability profiles across scientific settings, with model performance varying according to the procedures, objects, and interaction patterns involved in each discipline.

\paragraph{Performance by Experiment Viewpoint.}
Table~\ref{tab:viewpoint_breakdown} reveals distinct performance patterns across the two experiment viewpoints. Nearby-observer tasks account for 137 of the \numtasks{} tasks, while the remaining 67 are first-person tasks. On nearby-observer tasks, MiniMax~H3 ranks first with 52.2, followed by Seedance~2.0 at 47.1 and Wan~3.0 at 44.5. On first-person tasks, Wan~3.0 instead ranks first with 44.2, followed by Seedance~2.0 at 41.0 and MiniMax~H3 at 34.3. Thus, Seedance~2.0 remains among the top two models under both viewpoints, whereas the relative positions of MiniMax~H3 and Wan~3.0 differ substantially across recording perspectives.

The level-wise results further show that these viewpoint subsets are not uniformly ordered by task level. For first-person tasks, Seedance~2.0 and Wan~3.0 obtain higher scores on Level~2 than on Level~1 (46.0 vs.\ 34.7 and 46.6 vs.\ 41.3, respectively). The six open-weight models show a mixed pattern: Wan~2.2, Cosmos3~Super, and MiniMax~H3 score higher on Level~1, whereas LTX~2.3, Cosmos3~Nano, and LingBot~Video score higher on Level~2. For nearby-observer tasks, all six open-weight models score higher on Level~1 than on Level~2; among the commercial models, Seedance~2.0 scores higher on Level~2, whereas Wan~3.0 scores higher on Level~1. Because the task levels are independently curated, these contrasts are descriptive and should not be interpreted as isolated effects of task level or procedure length alone.

\subsubsection{Effect of Re-sampling}
\label{sec:resampling}
To assess the effect of sampling randomness, we conduct a best-of-three analysis on MiniMax~H3, Seedance~2.0, and Wan~3.0 using a fixed 40-task subset balanced across the two task levels. For each model, we generate two additional samples per task and compare the original Overall score with the score obtained by selecting the best of the three samples independently for each task. We also report the fraction of subset tasks for which the original generation remains a best-performing sample.

As shown in Table~\ref{tab:best_of_three}, the three models differ substantially in their sensitivity to re-sampling. MiniMax~H3 improves only slightly, from 55.3 to 57.8, with the original sample remaining best for 95\% of the tasks. In contrast, Seedance~2.0 and Wan~3.0 improve by 14.0 and 11.1 points, respectively, and both achieve higher best-of-three scores than MiniMax~H3. These results show that the ordering within this leading group is not robust to sampling variation. Although MiniMax~H3 obtains the highest Pooled Overall score under the single-sample protocol, Seedance~2.0 and Wan~3.0 exhibit substantially larger gains when additional samples are available. We therefore treat the main-table ordering as specific to the single-sample setting rather than as evidence that MiniMax~H3 consistently outperforms the other two generators across repeated sampling.

\subsubsection{Failure Cases and Attribution}
\label{sec:app_failure_attribution}
\begin{figure}[p]
    \centering
    \includegraphics[
        width=\linewidth,
        height=0.82\textheight,
        keepaspectratio
    ]{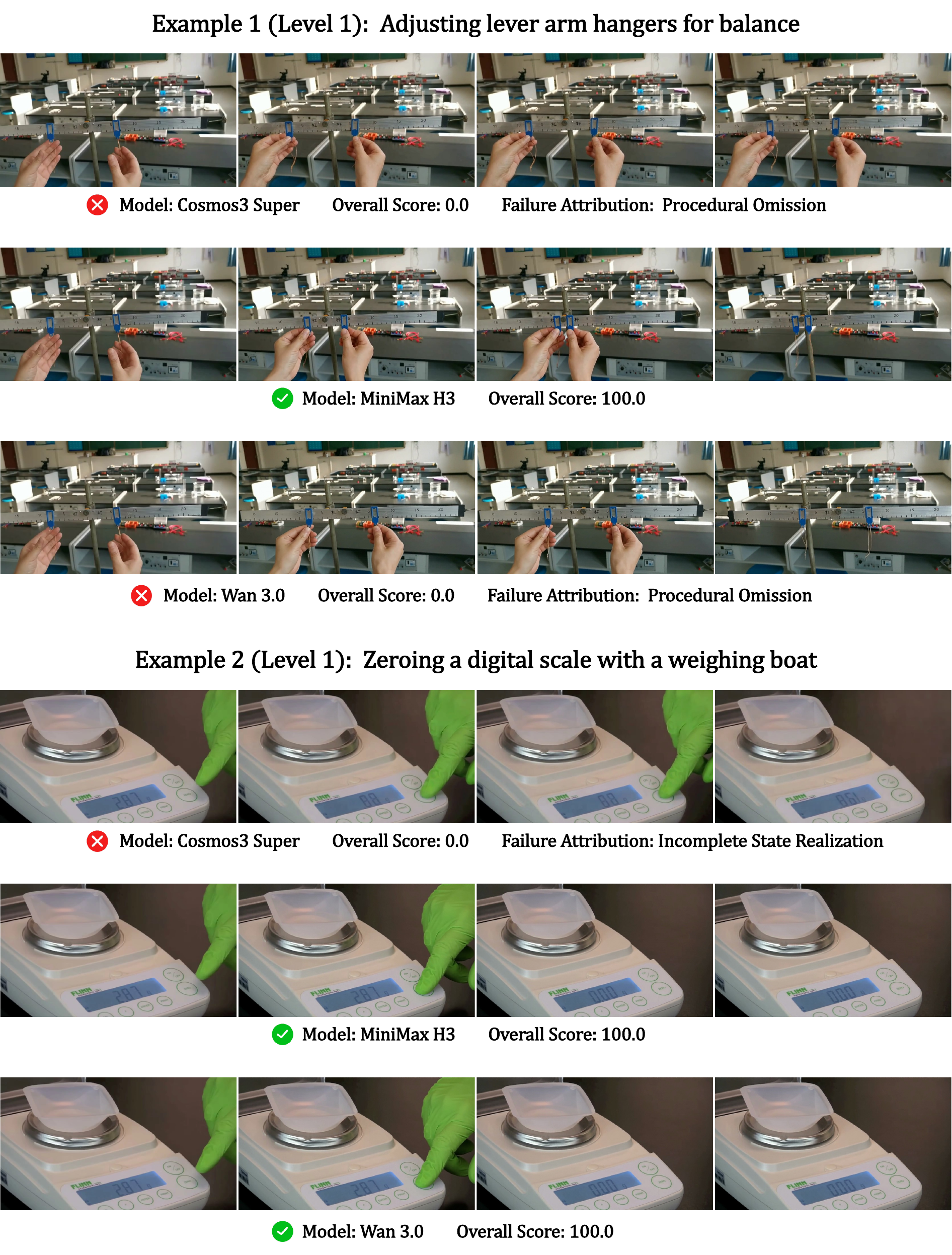}
    \caption{Videos generated by different models for two Level~1 example tasks.}
    \label{fig:example12}
\end{figure}

\begin{figure}[p]
    \centering
    \includegraphics[
        width=\linewidth,
        height=0.82\textheight,
        keepaspectratio
    ]{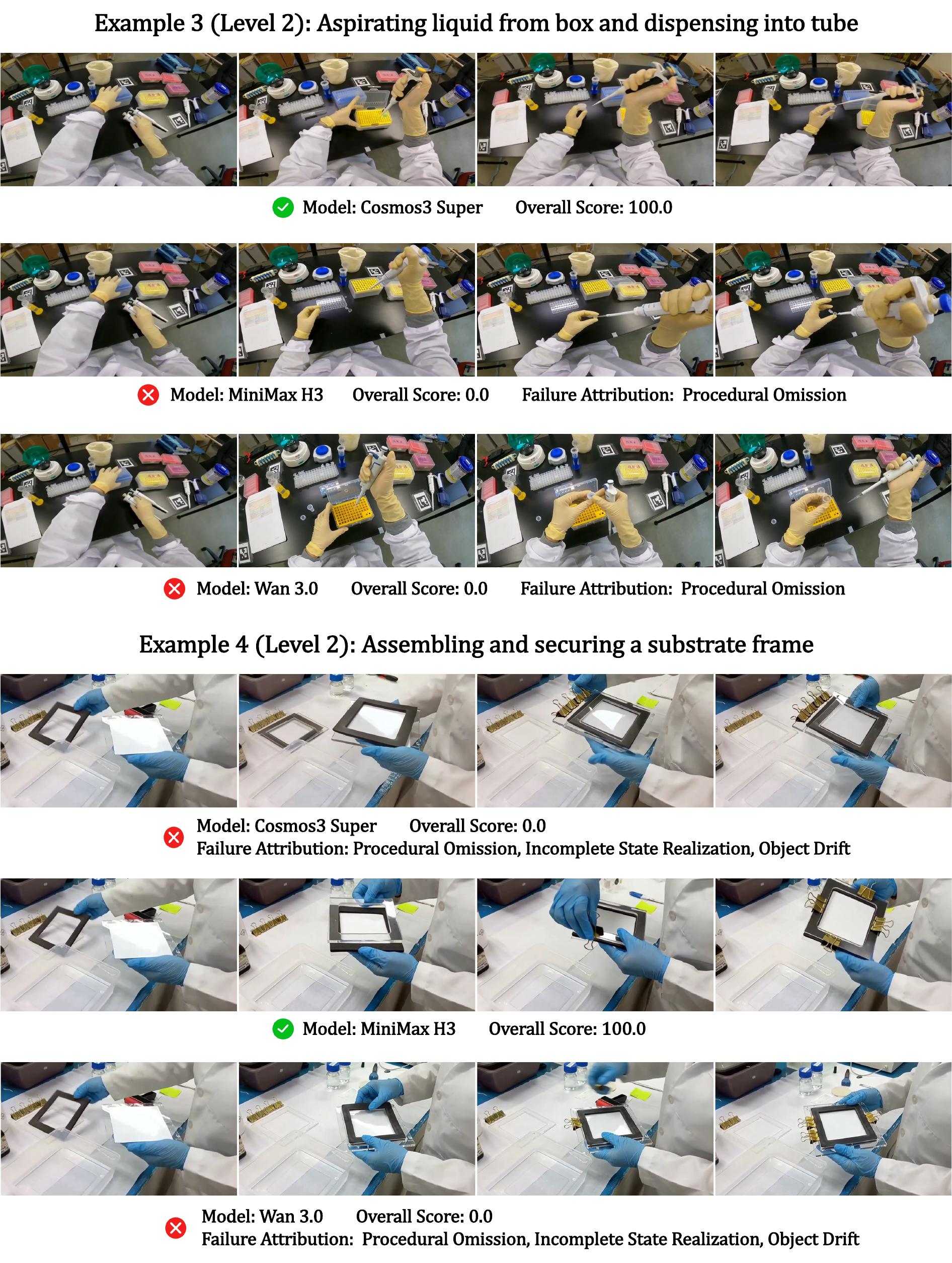}
    \caption{Videos generated by different models for two Level~2 example tasks.}
    \label{fig:example34}
\end{figure}

Figures~\ref{fig:example12} and~\ref{fig:example34} show videos generated by different models on four representative tasks across Level~1 and Level~2 settings. The Level~1 examples highlight two distinct requirements for successful task execution: correct action realization and correct state transition. In the lever-arm task, MiniMax~H3 completes the required inward adjustment, whereas Cosmos3 Super and Wan~3.0 exhibit procedural errors in the shown samples. In the scale-zeroing task, MiniMax~H3 and Wan~3.0 reach the target display state, while Cosmos3 Super fails to realize the expected zeroed state. These examples illustrate that even simple manipulation tasks require precise grounding of both the prescribed action and its intended outcome.

The Level~2 examples further emphasize the challenge of maintaining procedural and state consistency over multi-step interactions. In the pipetting task, MiniMax~H3 and Wan~3.0 omit different critical actions, while Cosmos3 Super completes the demonstrated procedure. In the substrate-frame assembly task, Cosmos3 Super and Wan~3.0 show a combination of procedural omission, incomplete state realization, and object drift, leading to incorrect final configurations. These cases suggest that complex manipulation requires not only locally correct actions but also consistent object states and spatial relations throughout the full execution sequence. The examples illustrate representative success and failure modes and are not intended to establish a ranking among generators.

\clearpage

\clearpage

\clearpage

\section{Task Schema and Prompt Templates}
\label{sec:task_prompts}

This section presents condensed versions of the task schema and prompts used to draft annotations, render generation instructions, construct task-conditioned QA items, and obtain structured VLM judgments. We retain the roles, inputs, principal constraints, and output contracts required to understand the pipeline while omitting repeated explanations and illustrative examples. The complete verbatim templates are provided in the accompanying project repository.

\subsection{Structured Task Schema}
\label{sec:task_schema}
For readability, we render the JSON task specification as semantically equivalent YAML.

\promptinput{prompts/task_schema.txt}

\subsection{Annotation Prompt for Level~1 Tasks}
\label{sec:level1_annotation_prompt}
\promptinput{prompts/l1_annotation_prompt.txt}

\subsection{Annotation Prompt for Level~2 Tasks}
\label{sec:level2_annotation_prompt}

\promptinput{prompts/l2_annotation_prompt.txt}

\subsection{Instruction Rendering Prompt for Level~1 Tasks}
\label{sec:level1_instruction_rendering_prompt}

\promptinput{prompts/l1_rendering_prompt.txt}

\subsection{Instruction Rendering Prompt for Level~2 Tasks}
\label{sec:level2_instruction_rendering_prompt}

\promptinput{prompts/l2_rendering_prompt.txt}

\subsection{Task-Conditioned QA Generation Prompt}
\label{sec:qa_generation_prompt}

\promptinput{prompts/qa_generation.txt}

\subsection{VLM Evaluation Prompt}
\label{sec:vlm_evaluation_prompt}

\promptinput{prompts/vlm_evaluation.txt}

\section{Limitations and Ethical Considerations}
\label{sec:limitations}

\paragraph{Dataset Scale and Coverage.}
Publicly available laboratory videos that satisfy our curation criteria remain limited. Most web videos contain shot transitions, subtitles, branding, or other visual overlays, whereas continuous, unedited recordings of a single laboratory procedure are comparatively rare. After screening, 81 source videos yield \numtasks{} short, safety-screened task clips, including \numlevelone{} Level~1 and \numleveltwo{} Level~2 tasks, spanning \numdisciplines{} disciplines. Consequently, \benchmark\ represents a relatively clean and controlled subset of real-world laboratory video rather than the full diversity of scientific procedures, environments, and equipment.

The source distribution is also imbalanced. Of the 81 source videos, 61 originate from Bilibili, accounting for 163 of the \numtasks{} task clips, while the professionally recorded ExpVid subset contributes 3 videos and 14 clips. Model performance may therefore be influenced in part by platform-specific recording styles, equipment, and presentation conventions. Coverage across disciplines and operation types is similarly uneven, ranging from 73 Physics tasks to 13 Materials Science tasks, and from 60 instrument-operation tasks to 7 heating/cooling tasks. Future extensions should increase source diversity and broaden the laboratory settings, procedure lengths, equipment, operation types, and scientific domains represented in the benchmark.

\paragraph{Sampling Variance.}
The main benchmark evaluates one generated sample for each model--task pair under fixed, model-specific generation settings, and therefore characterizes performance under a standardized run rather than each model's full sampling distribution. We partially examine this limitation through a targeted best-of-three analysis on a 40-task subset balanced across the two levels for the three leading generators. The original generation remains a best-performing sample for 80--95\% of the selected tasks, although additional sampling yields model-dependent gains concentrated in a minority of cases. This analysis shows that fine-grained ordering within the leading group can change under repeated sampling, but it does not characterize sampling variance for the other models or the full task set. Future work should evaluate multiple samples per task across all models to estimate complete generation distributions and ranking uncertainty.

\paragraph{Training-Data Contamination.}
Because the source videos are drawn from public platforms and existing research datasets, some benchmark content may have been included in the pretraining data of the evaluated generators. This possibility cannot be reliably audited for commercial models whose training corpora are not publicly disclosed. We therefore do not claim that \benchmark\ is free from training-data contamination. The paired reference videos provide evidence for evaluating whether a generated clip matches the target procedure, but they do not guarantee that the underlying visual content or procedure was unseen during model training.

\paragraph{Safety and Responsible Use.}
Generated laboratory videos should not be interpreted as operational instructions or as evidence that a procedure is safe to perform. They are not a substitute for expert supervision, institutional training, formal risk assessment, standard operating procedures, or other laboratory safety documentation. Our visual safety gate identifies only safety-relevant issues that are observable in the generated clip and cannot determine whether a procedure is safe in practice. Although we exclude hazardous procedures from the benchmark and report safety validity explicitly, benchmark performance does not constitute certification or endorsement of any generated procedure for real-world execution without independent expert review.

\section{Other Details}
\label{sec:other_details}

\subsection{Licenses and Terms of Use}
We evaluate commercial video-generation APIs and openly released model weights. For commercial APIs, we follow the corresponding provider terms of service and use generated outputs solely for research evaluation. For open-weight models and supporting tools, we use them under their respective licenses and cite the original authors. We do not redistribute third-party model weights or proprietary API outputs except where permitted by the applicable licenses or terms of service.

\subsection{Source-Video Provenance and Distribution}
For each source video, we record the corresponding URL or dataset identifier, source identifier, and discipline, and preserve this provenance for all derived task clips. We additionally retain the exact temporal boundaries of each derived task clip, which identify the corresponding conditioning frame and reference execution. To respect the terms and licenses of the original platforms and datasets, we plan to only release the benchmark task specifications, generation instructions, checklist annotations, evaluation code, and provenance metadata needed to identify the corresponding source material and reconstruct benchmark instances where the original sources remain accessible, under a research-friendly license. We will also make reasonable efforts to assist researchers with source-access or benchmark-reconstruction issues.

\subsection{Intended Use}
\benchmark\ is intended for academic research and diagnostic evaluation of procedural fidelity, visual consistency, and related reasoning capabilities in video generation models. The evaluated generators, VLM-based judges, and supporting libraries are used solely for research evaluation and data processing, and not for commercial training, individual profiling, or consequential real-world decision-making. Human annotations are used only to validate the reliability of VLM-as-judge evaluation, and are anonymized prior to analysis.

\subsection{Human Annotation: Recruitment, Consent, and Ethics Review}
All human annotations were produced by the paper's co-authors as part of the research process. No external participants were recruited through crowdsourcing platforms, participant pools, or other recruitment channels, and no separate compensation was provided. All annotators were informed that their judgments would support benchmark validation, including the assessment of VLM-as-judge reliability and evaluator calibration.

We did not collect names of annotators, contact information, demographic attributes, or other sensitive personal information for the purpose of this study. Annotations are used exclusively for benchmark validation and are reported only in aggregate. Because the annotation process involved only the paper's co-authors and did not recruit external human participants, we did not obtain separate institutional ethics review board approval or an exemption determination.

\clearpage

\end{document}